\documentclass[10pt,journal,compsoc]{IEEEtran}
\ifCLASSOPTIONcompsoc
  \usepackage[nocompress]{cite}
\else
  \usepackage{cite}
\fi
\ifCLASSINFOpdf
\else
\fi

\usepackage{graphicx}
\usepackage{mathptmx} 

\usepackage{latexsym}
\usepackage{times}
\usepackage{epsfig}
\usepackage[misc]{ifsym}
\usepackage{amsmath}
\usepackage{amssymb}
\usepackage{float}
\usepackage{cuted}
\usepackage{color}

\usepackage[ruled,vlined]{algorithm2e}
\usepackage{multirow}
\usepackage{caption}
\usepackage[colorlinks,linkcolor=blue]{hyperref}
\usepackage{ragged2e} 
\newcommand{\onedot}{\ifx\@let@token.\else.\null\fi\xspace}

\newcommand{\eg}{\emph{e.g}\onedot}
\newcommand{\ie}{\emph{i.e}\onedot}

\newcommand{\rev}[1]{{\color{black}#1}}

\begin{document}
%
\title{ViTPose++: Vision Transformer for Generic Body Pose Estimation}
%
%
%
%

\author{Yufei~Xu,~\IEEEmembership{Student Member,~IEEE},
        Jing~Zhang,~\IEEEmembership{Senior Member,~IEEE},
        Qiming~Zhang,~\IEEEmembership{Student Member,~IEEE},
        and~Dacheng~Tao,~\IEEEmembership{Fellow,~IEEE}
\thanks{
The project was supported by Australian Research Council Project FL-170100117.
Y. Xu and J. Zhang contributed equally to this paper. Corresponding author: Jing Zhang, Dacheng Tao.

Y. Xu, J. Zhang, Q. Zhang, and D. Tao are with the Sydney AI Centre and the School of Computer Science in the Faculty of Engineering at The University of Sydney, 6 Cleveland St, Darlington, NSW 2008, Australia (email: \{yuxu7116,qzha2506\}@uni.sydney.edu.au; \{jing.zhang1,dacheng.tao\}@sydney.edu.au).}
}

%
%

\markboth{Journal of \LaTeX\ Class Files,~Vol.~14, No.~8, August~2015}%
{Shell \MakeLowercase{\textit{et al.}}: Bare Demo of IEEEtran.cls for Computer Society Journals}
%



\IEEEtitleabstractindextext{%
\begin{abstract}
\justifying
Although no specific domain knowledge is considered in the design, plain vision transformers have shown excellent performance in visual recognition tasks. However, little effort has been made to reveal the potential of such simple structures for body pose estimation tasks. In this paper, we show the surprisingly good properties of plain vision transformers for body pose estimation from various aspects, namely simplicity in model structure, scalability in model size, flexibility in training paradigm, and transferability of knowledge between models, through a simple baseline model dubbed \textbf{ViTPose}. Specifically, ViTPose employs the plain and non-hierarchical vision transformer as an encoder to encode features and a lightweight decoder to decode body keypoints in either a top-down or a bottom-up manner. It can be scaled up from about 20M to 1B parameters by taking the advantage of the scalable model capacity and high parallelism of the vision transformer, setting a new Pareto front for throughput and performance. Besides, ViTPose is very flexible regarding the attention type, input resolution, and pre-training and fine-tuning strategy. Based on the flexibility, a novel \textbf{ViTPose++} model is proposed to deal with heterogeneous body keypoint categories in different types of body pose estimation tasks via knowledge factorization, \ie, adopting task-agnostic and task-specific feed-forward networks in the transformer. We also empirically demonstrate that the knowledge of large ViTPose models can be easily transferred to small ones via a simple knowledge token. Experimental results show that our ViTPose model outperforms representative methods on the challenging MS COCO Human Keypoint Detection benchmark at both top-down and bottom-up settings. Specifically, our largest single model ViTPose-G with 1B parameters sets a new record on the MS COCO test set without model ensemble. Furthermore, our ViTPose++ model achieves state-of-the-art performance simultaneously on a series of body pose estimation tasks, including MS COCO, AI Challenger, OCHuman, MPII for human keypoint detection, COCO-Wholebody for whole-body keypoint detection, as well as AP-10K and APT-36K for animal keypoint detection, without sacrificing inference speed. The source code and models are available at \href{https://github.com/ViTAE-Transformer/ViTPose}{https://github.com/ViTAE-Transformer/ViTPose}.
\end{abstract}

\begin{IEEEkeywords}
Vision transformer, Pose estimation, Top-down, Bottom-up, Pre-training, Transfer learning, Multi-task learning
\end{IEEEkeywords}}

\maketitle

\IEEEdisplaynontitleabstractindextext

%
\IEEEpeerreviewmaketitle

\IEEEraisesectionheading{\section{Introduction}\label{sec:introduction}}

\begin{figure}[!ht]
    \centering
    \includegraphics[width=0.9\linewidth]{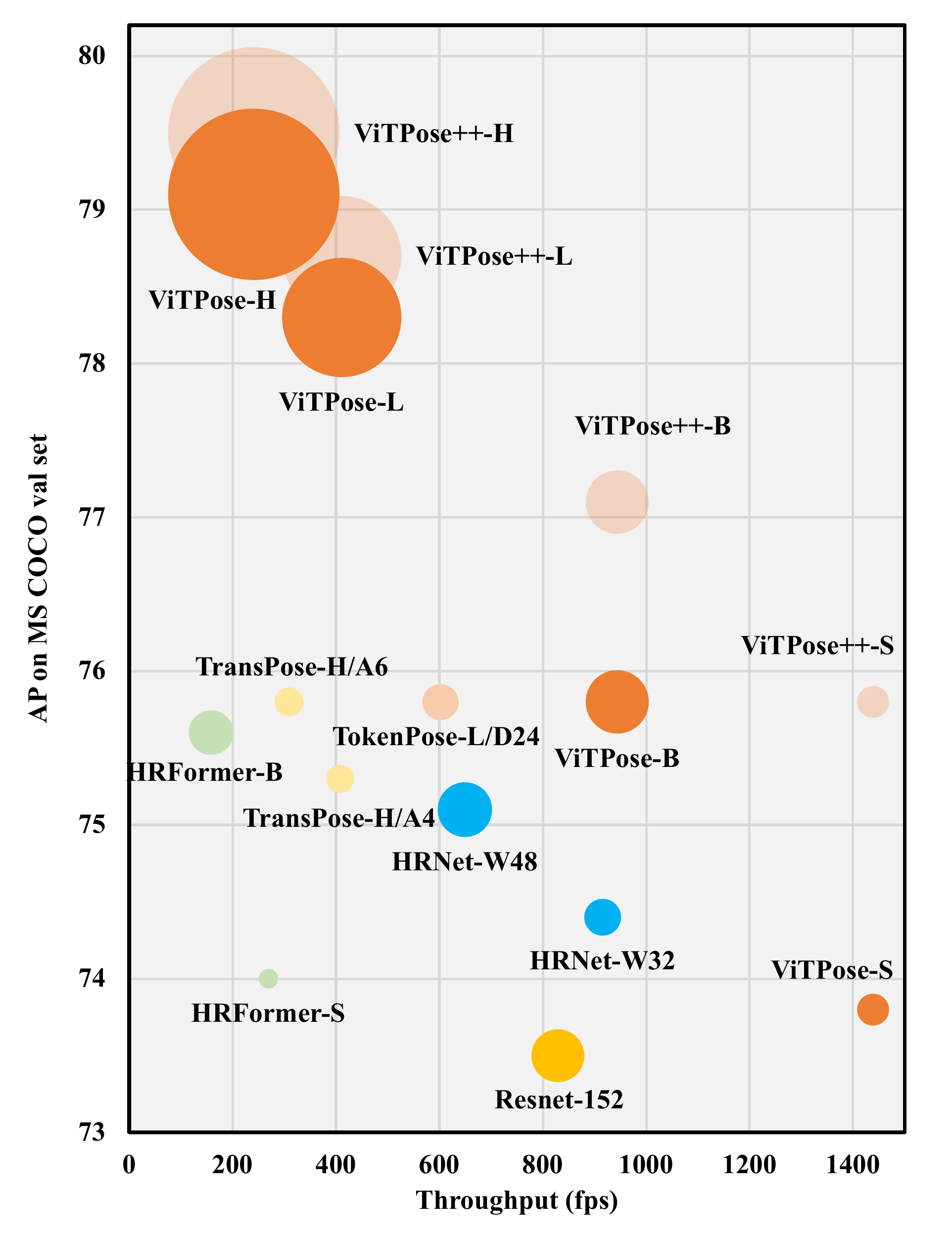}
    \caption{Comparison of ViTPose and SOTA methods on the MS COCO val set regarding model size, throughput, and accuracy. The size of each bubble represents the number of model parameters.}
    \label{fig:opening}
\end{figure}

\IEEEPARstart{B}{ody} keypoint detection (a.k.a. body pose estimation of human, animal, etc.) is one of the fundamental tasks in computer vision and has a wide range of real-world applications~\cite{zhang2020empowering,lin2020human,li2019crowdpose}. As a representative example, human pose estimation aims to localize human anatomical keypoints and is challenging due to the variations of occlusion, truncation, scales, and human appearances. To deal with these challenges, there has been rapid progress in deep learning-based methods~\cite{toshev2014deeppose,xiao2018simple,rafi2020self,zhang2020key}, which are typically built upon convolutional neural networks (CNN). 

Recently, vision transformers~\cite{dosovitskiy2020image,liu2021swin,carion2020end,strudel2021segmenter} have shown great potential in many computer vision tasks. Inspired by their success, different vision transformer structures have been explored for human pose estimation. Most of them adopt CNN as the backbone and then use a transformer of elaborate structures to refine the extracted features and model the relationship between the body keypoints. For example, PRTR~\cite{Li_2021_CVPR} incorporates both transformer encoders and decoders to gradually refine the locations of the estimated keypoints in a cascade manner. TokenPose~\cite{li2021tokenpose} and TransPose~\cite{yang2021transpose}, instead, adopt an encoder-only transformer structure to further process the features extracted by the CNN backbone. Inspired by the success of anchor-based methods in detection~\cite{carion2020end}, a query-based decoder is adopted in Poseur~\cite{mao2022poseur} with noisy augmentations. On the other hand, HRFormer~\cite{YuanFHLZCW21} employs the transformer to directly extract features and introduce high-resolution representations via multi-resolution parallel transformer modules. These methods have obtained superior performance on human pose estimation. However, they either need an extra CNN backbone for feature extraction or adopt transformer structures specially designed for pose estimation. Besides, most of them favor body pose estimation only on a single type, \ie, human, while ignoring the potential of transformers in modeling the correlations between different types of body keypoints (\eg, human and animal), which is essential for developing a foundation model for generic body keypoint detection. This motivates us to think from a different direction, \textit{how well can plain vision transformers do for body pose estimation towards a foundation model?}

To find the answer to this question, we propose a simple baseline model dubbed \textbf{ViTPose} and demonstrate the potential of simple plain vision transformers for human pose estimation in both the top-down and bottom-up manner. Then, a novel \textbf{ViTPose++} model is proposed to deal with multiple types of body keypoint detection via knowledge factorization. Specifically, ViTPose employs a plain and non-hierarchical vision transformer~\cite{dosovitskiy2020image} as the backbone to extract features for an input person instance or image, determined by the adopted pose estimation paradigm, \ie, either the top-down paradigm or the bottom-up paradigm. The backbone is pre-trained with masked image modeling pretext tasks, \eg, MAE~\cite{MaskedAutoencoders2021}, to provide a good initialization. Then, a lightweight decoder, which is composed of two deconvolution layers and one prediction layer, further processes the extracted features by up-sampling and regression to get the heatmaps of different keypoints. It is noteworthy that the associate maps are also regressed together with the heatmaps for the bottom-up paradigm. Despite no elaborate task-specific designs, ViTPose achieves a state-of-the-art (SOTA) performance of 80.9 AP on the challenging MS COCO test-dev set for human keypoint detection. ViTPose++ adopts the idea of the mixture of experts (MoE) in the backbone networks to factorize the knowledge between different types of body keypoints\footnote{We treat each type of body keypoint detection as an individual task.}, considering that different body pose estimation tasks may share common knowledge of body keypoints while also requiring task-specific knowledge. In detail, ViTPose++ decomposes the feed-forward networks (FFN) into shared and task-specific parts to encode the common and task-specific features, respectively. In this way, ViTPose++ effectively addresses the task conflict issue and thus achieves better performance on each task than the single-task and naive multi-task learning methods. It also sets new SOTA on four challenging benchmarks, including MS COCO~\cite{lin2014microsoft}, OCHuman~\cite{lin2020human}, MPII~\cite{andriluka20142d}, and AP-10K~\cite{yu2021ap}. Besides, ViTPose++ retains the inference speed as ViTPose since no more parameters and computations are introduced for each task.

Besides the superior performance, we also show the surprisingly good properties of ViTPose from various aspects, namely simplicity, scalability, flexibility, and transferability. 1) For simplicity, thanks to the strong feature representation ability of vision transformers, the ViTPose pipeline can be extremely simple. For example, it does not require any specific domain knowledge to design the backbone encoder and enjoys a plain and non-hierarchical encoder structure by simply stacking several transformer layers. The decoder can be further simplified to a single bilinear up-sampling layer followed by a common convolutional prediction layer with a negligible performance drop. This structural simplicity makes ViTPose enjoy better computational parallelism so that it reaches a new Pareto front for inference speed and performance, as shown in Fig.~\ref{fig:opening}. 2) In addition, the simplicity in structure brings the excellent scalability of ViTPose, which can benefit from the rapid development of scalable pre-trained vision transformers. Specifically, one can easily control the model size by stacking different numbers of transformer layers and setting different feature dimensions (\eg, using ViT-S, ViT-B, ViT-L, or ViT-H) to balance the inference speed and performance for various deployment requirements. 3) Furthermore, we demonstrate that ViTPose is very flexible in the training paradigm. It can adapt well to higher input and feature resolutions with minor modifications and deliver better pose estimation results. 
In addition, ViTPose can obtain competitive performance even if it is pre-trained using smaller unlabelled datasets or fine-tuned with the attention modules frozen at a lower training cost. 4) Last but not least, we show that the performance of small ViTPose models can be easily improved by transferring the knowledge from large ViTPose models through an extra learnable knowledge token, demonstrating a good transferability of ViTPose.

In summary, the main contribution of this paper is threefold. 1) We propose simple yet effective baseline models named ViTPose for human pose estimation and ViTPose++ for generic body keypoint detection. It obtains SOTA performance on representative benchmarks without elaborate designs for the backbone and pipeline. 2) The simple ViTPose model demonstrates to have surprisingly good properties, including structural simplicity, model size scalability, training paradigm flexibility, and knowledge transferability. These properties can shed light on the future development of vision transformer-based methods in the field of pose estimation.
3) Comprehensive experiments on challenging public benchmarks are conducted to evaluate and analyze the performance of ViTPose and ViTPose++, which set new SOTA on the benchmarks for different body keypoint detection tasks, making a solid step towards developing a foundation model for generic (body) keypoint detection.

\section{Related Work}

\subsection{Representative pose estimation methods}

\subsubsection{Top-down methods} 
{Pose estimation has undergone rapid development, transitioning from CNN-based approaches~\cite{xiao2018simple, rafi2016efficient} to the utilization of vision transformers~\cite{dosovitskiy2020image,YuanFHLZCW21,zhang2022vitaev2}. The majority of existing methods concentrate on estimating poses from provided human instances, following a top-down pipeline. Notably, G-MRI~\cite{papandreou2017towards} introduced Faster-RCNN for generating bounding boxes and proposed a combined regression and classification task for keypoint estimation. RMPE~\cite{fang2017rmpe} further employed a spatial transformer network to rectify noise in detected bounding boxes and facilitate subsequent pose estimation. To achieve accurate keypoint localization, some approaches employed high-resolution features in CNN networks through techniques such as skip feature concatenation~\cite{newell2016stacked} or highway structures~\cite{sun2019deep}. CPN~\cite{chen2018cascaded} introduced a two-stage network to refine estimated keypoints using hierarchical features. Other methods like OKDHP~\cite{li2021online} leveraged multiple-branch networks and dynamic fusion strategies to utilize multi-scale information effectively. AlphaPose~\cite{alphapose} improved performance by leveraging a symmetric keypoint localization technique. In contrast, SimpleBaseline~\cite{xiao2018simple} directly recovered high-resolution features with decoders. As vision transformers have demonstrated superior performance across various vision tasks~\cite{zhang2022segvit, li2022exploring}, some approaches explored the use of transformers as decoders after the CNN backbone~\cite{Li_2021_CVPR, li2021tokenpose, yang2021transpose}. For instance, TransPose~\cite{yang2021transpose} processed features extracted by CNN to model global relationships, while TokenPose~\cite{li2021tokenpose} introduced token-based representations and extra tokens to estimate the locations of occluded keypoints and model the relationship among keypoints. Poseur~\cite{mao2022poseur} explored an anchor-based strategy with noise augmentation using transformer decoders. These transformer-based pose estimation methods have achieved superior performance on popular human pose estimation benchmarks. However, these methods still rely on CNN backbones for feature extraction. In contrast, HRFormer~\cite{YuanFHLZCW21} utilized transformers to directly extract high-resolution features. It incorporated a parallel transformer module to gradually fuse multi-resolution features and achieved outstanding performance. Nevertheless, the extent to which attention mechanisms contribute to pose estimation remains unclear and few efforts have explored the potential of plain vision transformers for pose estimation. In this paper, we fill this gap by proposing a simple yet effective baseline model dubbed ViTPose based on the plain vision transformers.}

\subsubsection{Bottom-up methods} {In contrast to top-down methods that operate on individual human instances, bottom-up methods directly localize the keypoints of all humans present in the input images. DeepCut~\cite{pishchulin2016deepcut} and DeeperCut~\cite{insafutdinov2016deepercut} employ regression to estimate all keypoints and subsequently group them into separate individuals. OpenPose~\cite{openpose} adopts a two-branch network, with one branch dedicated to keypoint estimation and the other to grouping. Associate embedding~\cite{newell2017associative} simplifies the bottom-up pipeline by jointly estimating keypoint locations and associating them within the same network. Pifpaf~\cite{kreiss2019pifpaf} employs a Part Intensity Field and a Part Association Field to regress and associate body keypoints for all humans. SPM~\cite{nie2019single} simplifies the pipeline by directly predicting the root joint for each human and the displacement between the root joint and other keypoints based on a structured pose representation.
To enhance the performance of bottom-up methods, HigherHRNet~\cite{cheng2020higherhrnet} introduces multi-stage supervision. In contrast, we embark on the pioneering exploration of plain vision transformers for bottom-up pose estimation, which shows promising performance.}

\subsection{Vision transformer pre-training}
Inspired by the success of ViT~\cite{dosovitskiy2020image}, many different vision transformers~\cite{liu2021swin,xu2021vitae,wang2022crossformer,zhou2021elsa,wang2021pyramid,zhang2022vitaev2,wang2021scaled,zhang2022vsa} have been proposed, which are typically pre-trained on the ImageNet~\cite{deng2009imagenet} dataset at the fully supervised setting. However, the fully-supervised learning paradigm needs a large scale of labeled data, which is expensive. Besides, such pre-trained models may not generalize well on the tasks with a very different data distribution as ImageNet. To provide a better initialization for the vision transformers, self-supervised pre-training methods have been proposed, either in the contrastive way~\cite{chen2021empirical,xu2022regioncl} or the generative way~\cite{MaskedAutoencoders2021,bao2022beit}. A typical example is MAE~\cite{MaskedAutoencoders2021} that adopts a masked image modeling (MIM) pretext task inspired by the success of masked language modeling pretext tasks in natural language processing. Surprisingly, MIM pre-trained models on ImageNet without using the labels demonstrate better generalization performance on image classification and downstream tasks. In this paper, we focus on pose estimation tasks and adopt plain vision transformers with MIM pre-training as backbones. Besides, we explore whether or not using ImageNet for pre-training is necessary for pose estimation tasks. Surprisingly, using smaller unlabelled pose datasets for pre-training can also provide a good initialization.

\subsection{Foundation models}

Foundation model~\cite{bommasani2021opportunities}, with the aim to deal with diverse real-world problems with a unified model, has received increasing attention recently. Owing to their strong representation and scalability abilities, vision transformers have become popular in building foundation models and leveraging massive data from multiple modalities for training. For example, Florence~\cite{yuan2021florence} employs different adapter networks on the pre-trained backbones to handle many tasks, including classification, detection, and segmentation. Encoder-decoder structures are adopted in \cite{alayrac2022flamingo} to solve multi-modality tasks via a novel Perceiver module. To further enhance the representation ability of backbone networks, MoE~\cite{shazeer2017outrageously} is adopted by allowing the network to ensemble the outputs from different sub-networks without introducing too much extra computational costs. Typically, each MoE layer contains a gate layer to determine the appropriate expert to use and fuse the outputs of the selected experts in a weighted sum manner, \eg, UniPerceiver-MoE~\cite{zhu2022uni} adopts MoE-based attention layers and FFN layers to construct the networks. {In this paper, we aim to explore the potential of plain vision transformers for generic body keypoint detection. We propose ViTPose++, which decomposes the FFN layers into task-agnostic and task-specific experts to handle different types of body pose estimation tasks simultaneously.} Without extra parameters and computations, ViTPose++ outperforms ViTPose trained in either a single-task or multi-task manner and sets new SOTA on challenging public benchmarks.

\begin{figure*}[ht]
    \centering
    \includegraphics[width=0.9\linewidth]{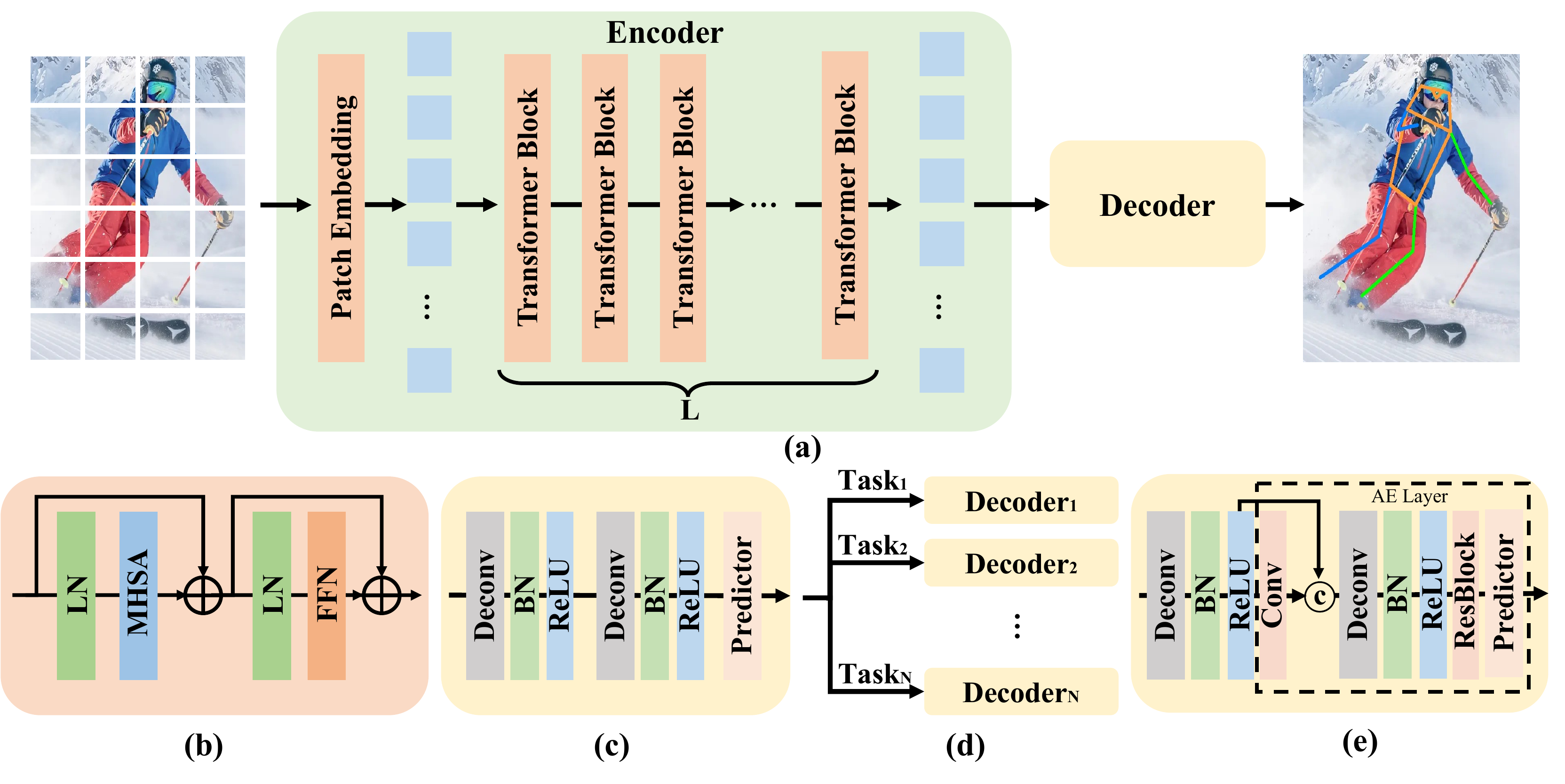}
    \caption{(a) The diagram of the proposed ViTPose model. (b) The transformer block. (c) The classic decoder in the top-down paradigm. {(d) The multi-task decoder used to deal with multiple datasets}. (e) The decoder in the bottom-up paradigm.}
    \label{fig:framework}
\end{figure*}

\subsection{Comparison to the conference version}

A preliminary version of this paper is presented in \cite{xu2022vitpose}. This paper extends the previous study with three major improvements. \textbf{1)} We explore more model sizes of ViTPose to thoroughly compare with representative models from many aspects, \ie, number of parameters, inference speed, input resolution, and performance. For example, the ViTPose-S model with 24M parameters obtains 73.8 AP with 1,432 fps on MS COCO, which is better than ResNet-50~\cite{xiao2018simple} (71.8 AP, 1,351 fps) and ResNet-152~\cite{xiao2018simple} (73.5 AP, 829 fps) and comparable with the frontier transformer-based model HRFormer-S~\cite{YuanFHLZCW21} (73.8 AP, 269 fps) but with a much faster speed. As demonstrated in Fig.~\ref{fig:opening}, different ViTPose variants from the small to large model sizes set a new Pareto front for throughput and performance, demonstrating the superiority of plain vision transformers for human pose estimation. \textbf{2)} We extend ViTPose to the bottom-up paradigm from the top-down paradigm by directly predicting the keypoint locations and their associations. The experimental results on the MS COCO dataset again demonstrate the potential and flexibility of vision transformers, \ie, plain vision transformers can perform well for human pose estimation in both top-down and bottom-up manners. \textbf{3)} A novel model dubbed ViTPose++ is further proposed to deal with multiple types of body pose estimation tasks via knowledge factorization. It obtains better performance than previous representative methods on public challenging datasets, \ie, MS COCO~\cite{lin2014microsoft}, OCHuman~\cite{zhang2019pose2seg}, MPII~\cite{andriluka20142d}, AI Challenger (AIC)~\cite{wu2017ai}, COCO-Wholebody (COCO-W)~\cite{jin2020whole}, AP-10K~\cite{yu2021ap10k}, and APT-36K~\cite{yang2022apt}, for various types of body keypoint detection, making the first attempt towards developing a foundation model for generic (body) keypoint detection. Besides, more experiment results, ablation studies, and analyses are presented. We also provide some visual results to demonstrate the promising performance of ViTPose++.

\section{ViTPose}

As shown in Fig.~\ref{fig:framework}, ViTPose employs a plain vision transformer for feature extraction and is compatible with different decoders for keypoint estimation, \eg, using classic (c) or simple (d) decoder in top-down keypoint estimation and associate embedding layer (e) in a bottom-up manner. Different properties of ViTPose are illustrated, \ie, simplicity, scalability, flexibility, and transferability. By exploiting these properties, a novel ViTPose++ model is further proposed and obtains SOTA performance on various pose estimation datasets. The details will be described below.

\subsection{The simplicity of ViTPose}
\label{subsec:structuresimplicity}
The goal of this paper is to provide a simple yet effective vision transformer baseline for body pose estimation tasks and explore the potential of plain and non-hierarchical vision transformers~\cite{dosovitskiy2020image}. Thus, we keep the structure as simple as possible and try to avoid fancy but complex modules, even though they may improve performance. To this end, we simply append several decoder layers after the transformer backbone to regress the heatmaps of keypoints, as shown in Fig.~\ref{fig:framework}(a).
Specifically, given a person instance image $X \in \mathcal{R}^{{H} \times {W} \times 3}$, it is first embedded into tokens via a patch embedding layer, \ie, $F \in \mathcal{R}^{\frac{H}{d} \times \frac{W}{d} \times C}$, where $H$ and $W$ are the height and width of the input image, respectively, $d$ (\eg, 16 by default) is the down-sampling ratio of the patch embedding layer, and $C$ is the channel dimension. Then, the tokens are processed by several transformer layers, each of which is consisted of a multi-head self-attention (MHSA) layer and an FFN, \ie, 
\begin{equation}
\begin{aligned}
    F_{i+1}^{'} &= F_{i} + {\rm MHSA}({\rm LN}(F_{i})), \\
    F_{i+1} &= F_{i+1}^{'} + {\rm FFN}({\rm LN}(F_{i+1}^{'})),
\end{aligned}
\end{equation}
where $F_i$ denotes the output of the $i$th transformer layer and we use $F_0={\rm PatchEmbed}(X)$ to denote the features after the patch embedding layer. It should be noted that the spatial and channel dimensions are constant for each transformer layer. We denote the output feature of the backbone network as $F_{out} \in \mathcal{R}^{\frac{H}{d} \times \frac{W}{d} \times C}$. 

\begin{figure}
    \centering
    \includegraphics[width=0.9\linewidth]{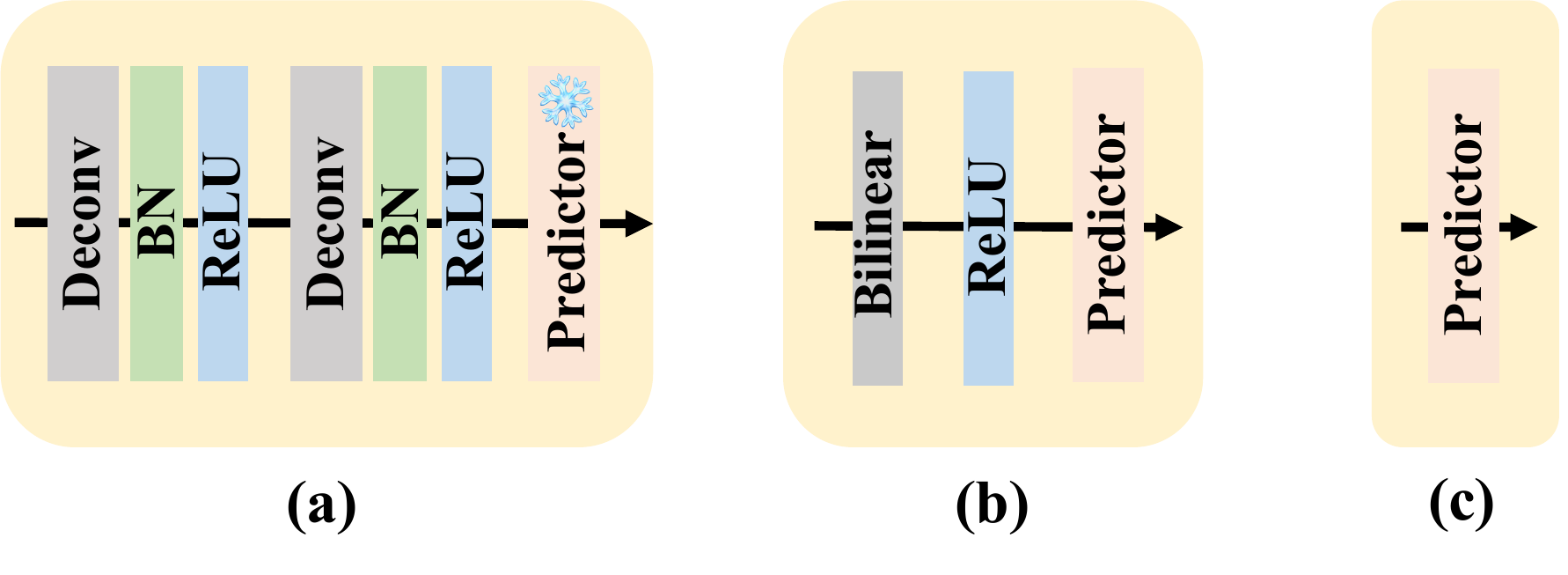}
    \caption{{(a) Classic decoder (Fig.~\ref{fig:framework}(c)) with a frozen predictor. (b) Simple decoder. (c) Minimal decoder.}}
    \label{fig:decoderDesign}
\end{figure}

{We adopt different decoder designs to process the features extracted from the backbone network and regress the heatmaps, \ie, the classic decoder, the simple decoder, and the minimal decoder, as shown in Fig.~\ref{fig:framework}(c) and Fig.~\ref{fig:decoderDesign}.} The classic decoder is composed of two deconvolution blocks, each containing one deconvolution layer followed by batch normalization~\cite{ioffe2015batch} and ReLU~\cite{agarap2018deep}. Following the common settings in previous methods~\cite{xiao2018simple,zhang2021towards}, each deconvolution blocks first up-samples the feature maps by a factor of 2. Then, an $1 \times 1$ convolution prediction layer is used to regress the heatmaps, \ie,
\begin{equation}
    K = {\rm Conv}_{1 \times 1} ({\rm Deconv}({\rm Deconv}(F_{out}))),
\end{equation}
where $K \in \mathcal{R}^{\frac{H}{4} \times \frac{W}{4} \times N_k}$ denotes the heatmaps of keypoints and $N_k$ is the number of keypoints, \eg, 17 for the MS COCO dataset.

{In addition to the classic decoder, we propose three simplified designs: classic-FP, simple decoder, and minimal decoder, to leverage the strong representation capabilities of the vision transformer backbone. Classic-FP retains the same structure as the classic decoder but with a randomly initialized and frozen predictor layer. For the simple decoder as shown in Fig.~\ref{fig:decoderDesign}(b), the feature maps extracted by the backbone are upsampled by a factor of 4 using bilinear interpolation and then fed into a ReLU activation function and a $3 \times 3$ convolutional layer to predict the heatmaps, \ie,
\begin{equation}
    K = {\rm Conv}_{3 \times 3} ({\rm Bilinear} ({\rm ReLU}(F_{out}))).
\end{equation}

The minimal decoder is the most simple design, which utilizes only a single linear projection layer for heatmap prediction, \ie,
\begin{equation}
    \begin{aligned}
        K' &= {\rm Conv}_{1 \times 1} (F_{out}) \in \mathcal{R}^{\frac{H}{16} \times \frac{W}{16} \times (16N_k)}, \\
    K &= {\rm PixelUnshuffle}(K') \in \mathcal{R}^{\frac{H}{4} \times \frac{W}{4} \times N_k},
    \end{aligned}
\end{equation}
where PixelUnshuffle~\cite{shi2016real} refers to the efficient operator that restores the spatial dimensions of the heatmap by reshaping it from the channel dimension to the spatial dimension. 
This design choice allows for a focused investigation into the representation capability of vision transformers in pose estimation}

\subsection{The scalability of ViTPose}
{While small-scale CNN-based models have shown promising results, they often face challenges when scaling up due to the overfitting problem caused by a strong inductive bias, as discussed in \cite{tolstikhin2021mlp}. One intriguing characteristic of vision transformers is their improved performance as the model size increases~\cite{dosovitskiy2020image}. However, it remains under-explored whether this scalability holds true for specific downstream tasks such as pose estimation. Based on ViTPose, we can easily control its model size by adjusting the number of stacked transformer layers and feature dimensions. This allows us to develop different sizes of models and fully exploit the benefits of pre-trained plain vision transformers. 
Specifically, we use the vision transformers of different model sizes as the backbone, \ie, ViT-B, ViT-L, ViT-H~\cite{dosovitskiy2020image}, and ViTAE-G~\cite{zhang2022vitaev2}, which are MIM pre-trained on ImageNet, and fine-tune them on the MS COCO dataset to investigate the scalability of ViTPose.} For ViT-H and ViTAE-G, which use patch embedding with size $14 \times 14$ during pre-training, we use zero padding to reformulate a patch embedding with size $16 \times 16$ for the same fine-tuning setting with ViT-B and ViT-L. As shown in Fig.~\ref{fig:opening}, ViTPose delivers continuing performance gains with the increased model size.

\subsection{The flexibility of ViTPose}
\label{subsec:flexibility}
\textbf{Pre-training data.} Pre-training the backbone networks on ImageNet~\cite{deng2009imagenet} has been a \textit{de facto} routine for a good initialization of network parameters. {However, this approach necessitates the use of large amounts of additional data beyond the pose data, thereby imposing higher requirements, particularly for vision transformers employed in pose estimation tasks. This motivates us to investigate whether it is possible to alleviate the data requirements for vision transformers by exclusively utilizing pose data throughout the entire training phase.} To answer this question, apart from the default setting of ImageNet~\cite{deng2009imagenet} pre-training, we also use MAE pre-training on MS COCO~\cite{lin2014microsoft} and a combination of MS COCO and AIC~\cite{wu2017ai}, respectively, by random masking 75\% patches from the images and reconstructing those masked patches. Then, we use the pre-trained weights to initialize the backbone of ViTPose and fine-tune it on the MS COCO dataset. Surprisingly, although the volume of the pose data is much smaller than ImageNet, ViTPose trained only with pose data can obtain competitive performance, demonstrating its flexibility regarding the pre-training data.

\textbf{Resolution of images and features.} We vary the input image size and down-sampling ratios $d$ of ViTPose to evaluate its flexibility regarding the input and feature resolution. Specifically, to adapt ViTPose to input images at a higher resolution, we simply resize the input images and train the model on them accordingly. Besides, to adapt the model to lower down-sampling ratios, \ie, higher resolution of feature maps, we simply change the stride of the patch embedding layer to partition tokens with overlap while keeping the size of each patch. We show that the performance of ViTPose increases consistently regarding either higher input resolution or higher feature resolution.

\textbf{Attention type.} Using full attention on high-resolution feature maps will cause a huge memory footprint and computational cost due to the quadratic computational complexity of vanilla self-attention. Window-based attention with relative position embedding~\cite{li2022exploring,li2021improved} have been explored to address this issue and can be used to deal with high-resolution feature maps. However, simply using window-based attention for all transformer blocks degrades the performance due to the lack of global context modeling ability. To address the problem, we adopt two techniques. \ie, 1) \textit{Shift window:} Instead of using fixed windows for attention calculation, we use the shifted window mechanism~\cite{liu2021swin} to help broadcast the information between adjacent windows. 2) \textit{Pooling window:} Apart from the shifted window, we try another solution via pooling. Specifically, we first extract the global token for each window by average pooling all the tokens in the window. The global tokens from all windows are then appended to the tokens in each window to serve as key and value tokens for attention calculation, thus enabling cross-window information exchange. In addition, we further show that the two techniques are complementary and can work together to improve performance and reduce memory footprint without introducing extra parameters.

\textbf{Fine-tuning strategy.} As demonstrated in NLP fields~\cite{liu2021pre,alayrac2022flamingo}, pre-trained transformer models can generalize well to other tasks by only tuning partial parameters. To investigate whether it still holds for vision transformers, we fine-tune ViTPose on MS COCO with all parameters unfrozen, MHSA frozen, and FFN frozen, respectively. We empirically demonstrate that with the MHSA frozen, ViTPose obtains comparable performance to the fully fine-tuning setting while using less memory footprint.

\textbf{Keypoint detection paradigm.} Although some works~\cite{YuanFHLZCW21,li2021tokenpose,mao2022poseur} have explored vision transformers for body keypoint detection in the top-down paradigm, it still needs to be determined how well the vision transformers, especially the plain vision transformers, can perform in the bottom-up paradigm. To this end, we thoroughly explore the potential of plain vision transformers in both top-down and bottom-up paradigms. Technically, in the top-down paradigm, the individual human instance is fed into the backbone network for feature extraction. In contrast, the whole image with several human instances is used as input to the backbone network in the bottom-up paradigm. The associate embedding technique~\cite{newell2017associative} is adopted to predict the keypoints and tag vectors based on the extracted features. Instead of using 1/16 feature resolution and vanilla self-attention in the top-down paradigm, we adopt 1/8 feature resolution and window-based attention in the bottom-up paradigm. The decoder of HigherHRNet~\cite{cheng2020higherhrnet} is adopted with an extra up-sample layer as shown in Fig.~\ref{fig:framework} (d), \ie,
\begin{equation}
    K = AE(Deconv(F_{out})),
\end{equation}
where $AE$ represents the associate embedding layer used to predict the tag vectors and the keypoint locations.

\subsection{The transferability of ViTPose}
One common method to improve the performance of small models is to transfer the knowledge from larger ones, \ie, knowledge distillation~\cite{hinton2015distilling,gou2021knowledge}. {Some interesting distillation methods~\cite{li2021online} have been developed to enhance the performance of small-scale models in pose estimation tasks. In our study, we show that even a very simple distillation method can effectively improve the performance of ViTPose.}
Specifically, given a teacher network $T$ and a student network $S$, a simple distillation method is to add an output distillation loss $L_{t \to s}^{od}$ to force the student network's output imitating the teacher network's output, \eg, 
\begin{equation}
    L_{t \to s}^{od} = {\rm MSE}(K_s, K_t),
\end{equation}
where $K_s$ and $K_t$ are the outputs (\eg, heatmaps) from the student and teacher network given the same input. 

{Transformers exhibit a distinct property of flexibility in handling inputs of various sizes. For instance, it has been shown that attaching several learnable prompt tokens to the inputs can improve transformers' performance on specific tasks~\cite{liu2022p}. These learnable tokens effectively capture task-related information from the training data, aiding in the optimization of transformers. Motivated by this observation, we propose a simple token-based distillation method to bridge the gap between large and small models. It complements the heatmap-based distillation methods and further enhances the performance of the models.} Specifically, we randomly initialize an extra learnable knowledge token $t$ and append it to the visual tokens after the patch embedding layer of the teacher model. Then, we freeze the well-trained teacher model and only update the knowledge token, \ie,
\begin{equation}
    t^{*} = \mathop{\arg\min}_{t}({\rm MSE} (T (\{t; X\}), K_{gt}),
\end{equation}
where $K_{gt}$ is the ground truth heatmaps, $X$ is the input images, $T(\{t; X\})$ denotes the predictions of the teacher, and $t^{*}$ represents the optimal token that minimizes the loss. Then, the knowledge token $t^{*}$ is frozen and concatenated with the visual tokens in the student network during training, thus transferring the knowledge from teacher to student. The loss for the student network is:
\begin{equation}
L_{t \to s}^{td} = {\rm MSE}(S(\{t^{*};X\}), K_{gt}),
\end{equation}
\begin{equation}
L_{t \to s}^{tod} = {\rm MSE}(S(\{t^{*};X\}), K_t) + {\rm MSE}(S(\{t^{*};X\}), K_{gt}),
\end{equation}
where $L_{t \to s}^{td}$ and $L_{t \to s}^{tod}$ denote the token distillation loss and its combination with the output distillation loss, respectively.

\subsection{ViTPose++}\label{subsec:methodMoE}

{A basic requirement for generic body keypoint detection should be able to deal with different body pose estimation tasks after training on multiple datasets with heterogeneous categories of body keypoint annotations.} One critical challenge is how to deal with the differences of body keypoints in different pose estimation tasks, \eg, 
the same keypoint (e.g., nose) of humans and animals with distinct appearances, and the different categories of keypoints in COCO-W~\cite{jin2020whole} which are absent in MS COCO~\cite{lin2014microsoft} and MPII~\cite{andriluka20142d}. Moreover, the data distribution of different species is also different, \eg, the human head is always above the shoulder, while the cow's head is always to the left or right of the shoulder.

\begin{figure}
    \centering
    \includegraphics[width=0.8\linewidth]{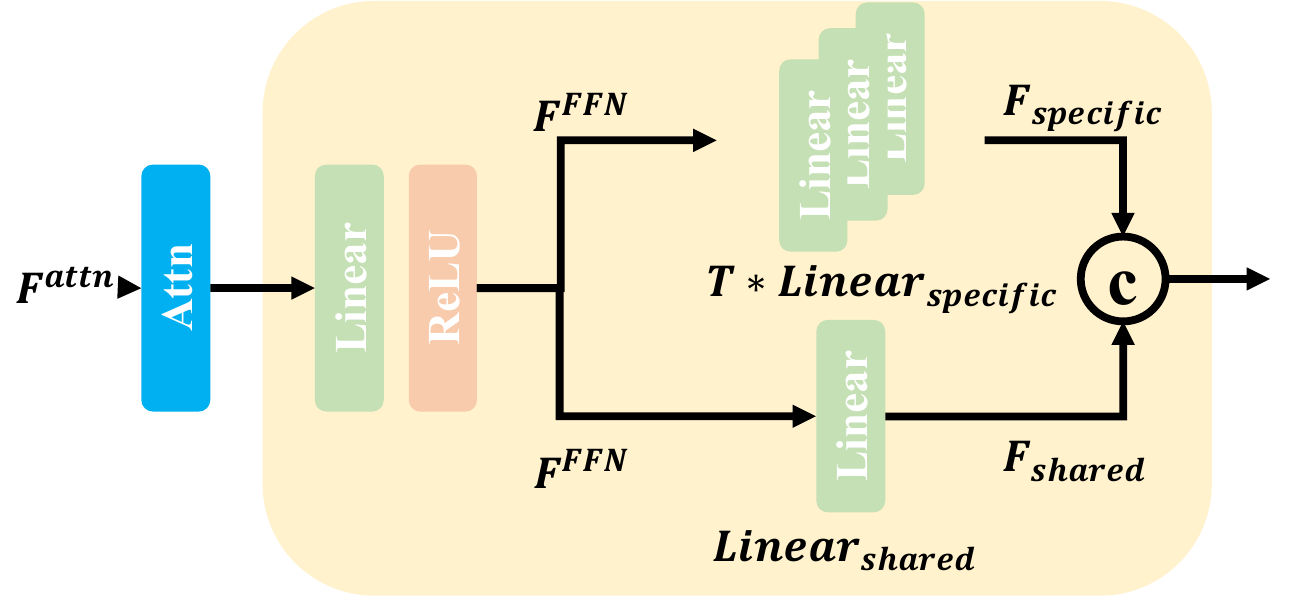}
    \caption{{The structure of FFN in the proposed ViTPose++ model. $T$ is the number of tasks processed by the model.}}
    \label{fig:moe}
\end{figure}

A naive solution is to train a ViTPose model by multi-task learning, \ie, using a shared backbone and different decoders, each of which is responsible for a specific type of pose estimation task. However, there may be conflicts between different tasks~\cite{liu2020towards}, affecting learning performance. In this paper, we propose a novel model dubbed ViTPose++ to address the challenge from the perspective of knowledge factorization. Specifically, since MHSA layers are not sensitive to pose estimation tasks (as evidenced by the experiment results of fine-tuning strategy in Section~\ref{sec:partial}), we adopt the idea of MoE~\cite{shazeer2017outrageously}, \ie, splitting FFN layers into a task-agnostic expert and multiple task-specific experts to encode the common and task-specific knowledge for pose estimation, respectively. Similar to the aforementioned naive multi-task learning method, we use a task-specific decoder for each type of body pose estimation task.
Technically, we take one transformer block as an example to illustrate the proposed ViTPose++. {As shown in Fig.~\ref{fig:moe},} given the output feature $F^{attn}$ of MHSA, it is processed by the first linear layer of FFN, which is shared by the MoE, \ie,
\begin{equation}
    F^{FFN} = ReLU(Linear(F^{attn})).
\end{equation}
Then, the output feature $F^{FFN} \in \mathcal{R}^{N \times \gamma C}$ is fed into separate linear layers (\ie, task-agnostic expert and task-specific expert), where $N$ represents the number of tokens and $\gamma$ is the expansion ratio of FFN, which is set to 4 by default. The two kinds of experts project $F^{FFN}$ to $F^{shared}$ and $F^{specific}$ with a channel dimension of $(1 - \alpha)C$ and $\alpha C$, respectively, \ie, 
\begin{equation}
    \begin{aligned}
        F^{shared} &= Linear_{shared}^{\gamma C \to (1 - \alpha) C} (F^{FFN}), \\
        F^{specific} &= Linear_{specific}^{\gamma C \to \alpha C} (F^{FFN}), \\
    \end{aligned}
\end{equation}
where $\alpha$ is the partition ratio used to balance the shared and task-specific experts and is set to 0.25 by default. Note that the parameters of the shared expert are trained using all the data, while the parameters of the task-specific expert are trained only using the data for the corresponding task. Then, $F^{shared}$ and $F^{specific}$ are concatenated along the channel dimension to form the output of the transformer block. Given an input image from the training set for a specific pose estimation task, after getting the encoded feature from the transformer backbone described above, it is fed into the corresponding decoder to regress the heatmaps. 

During inference for each type of pose estimation task, the shared and task-specific linear layers are merged into a single layer for parallel computation. In this way, ViTPose++ brings no extra parameters and computational costs compared with the ViTPose model while being able to serve as a foundation model for generic body pose estimation.

\begin{table*}[htbp]
  \centering
  \footnotesize
  \caption{Hyper-parameter settings for training ViTPose on the MS COCO dataset and the multiple datasets. The hyper-parameters before and after the slash are for ViTPose and ViTPose++, respectively.}
    {\begin{tabular}{cccccc}
    \hline
    Model & Batch Size & Learning rate & \multicolumn{1}{l}{Weight decay} & \multicolumn{1}{l}{Layer wise decay} & \multicolumn{1}{l}{Drop path rate} \\
    \hline
    ViTPose-S/ViTPose++-S & 512/1024 & 5e-4/1e-3 & 0.1   & 0.80   & 0.10 \\
    ViTPose-B/ViTPose++-B & 512/1024 & 5e-4/1e-3 & 0.1   & 0.75   & 0.30 \\
    ViTPose-L/ViTPose++-L & 512/1024 & 5e-4/1e-3 & 0.1   & 0.80   & 0.50 \\
    ViTPose-H/ViTPose++-H & 512/1024 & 5e-4/1e-3 & 0.1   & 0.80   & 0.55 \\
    ViTPose-G & 512 & 5e-4 & 0.1   & 0.85  & 0.55 \\
    \hline
    \end{tabular}}%
  \label{tab:implementation}%
\end{table*}%

\section{Experiments}
\subsection{Datasets and evaluation metrics}
\textbf{Datasets.} We adopt the MS COCO dataset to evaluate the performance of ViTPose and use several different pose estimation datasets to evaluate the proposed ViTPose++, \eg, MS COCO~\cite{lin2014microsoft}, AIC~\cite{wu2017ai}, and MPII~\cite{andriluka20142d} for human pose estimation, COCO-W~\cite{jin2020whole} for all body keypoint detection including those of face, hands, and feet, AP-10K~\cite{yu2021ap10k} and APT-36K~\cite{yang2022apt} for animal pose estimation. Interhand2.6M~\cite{moon2020interhand2} is also used to evaluate our models' data efficiency in transfer learning. OCHuman~\cite{zhang2019pose2seg} dataset is only used for evaluation to measure the performance of different models in dealing with occluded persons. Specifically, \textbf{MS COCO} contains 118K images and 150K human instances with at most 17 keypoint annotations for each instance. The dataset is under the CC-BY-4.0 license. \textbf{COCO-W} adopts the images from the MS COCO dataset but provides additional annotations on the face, feet, and hands of each person instance, resulting in at most 133 keypoints for each instance. \textbf{MPII} is under the BSD license and contains 25K images and over 40K human instances. There are at most 16 human keypoints for each instance annotated in this dataset. \textbf{AIC} is much bigger than other datasets and contains over 200K training images and 350 human instances, with at most 14 keypoints for each annotated instance, including shoulder, elbow, wrist, hip, knee, ankle, and head. \textbf{OCHuman} contains human instances with heavy occlusions and is only used for evaluation. It consists of 4K images and 8K instances. \textbf{AP-10K} dataset follows the CC-BY-4.0 license and is used for animal pose estimation. It contains 10K images with 54 different animal categories. Similarly, \textbf{APT-36K} dataset contains 36K images belonging to 30 different animal categories. 17 different keypoints are annotated in AP-10K and APT-36K datasets for each animal instance. \textbf{Interhand2.6M} contains 2.6M labeled hand frames generated from different subjects with both 2D and 3D annotations. We focus on 2D hand pose estimation temporally and leave 3D pose estimation~\cite{sigal2021human} as our future work.

\textbf{Metrics.} We adopt the average precision (AP) as the primary evaluation metric on most of the datasets, including COCO~\cite{lin2014microsoft}, AIC~\cite{wu2017ai}, COCO-W~\cite{jin2020whole}, OCHuman~\cite{zhang2019pose2seg}, and AP-10K~\cite{yu2021ap10k}, AP-36K~\cite{yang2022apt}. It is calculated by evaluating the object keypoint similarity (OKS) using different thresholds from 0.5 to 0.95~\cite{lin2014microsoft}. A loose metric AP$_{50}$ and a strict metric AP$_{75}$ are also utilized by setting the threshold to 0.5 and 0.75, respectively. PCKh is adopted as the evaluation metric on the MPII~\cite{andriluka20142d} dataset following the common practice, which evaluates the accuracy of each keypoint with a matching threshold related to the head segment length.

\begin{table*}[htbp]
  \centering
  \footnotesize
  \caption{Ablation study of the backbone and decoder in ViTPose on the MS COCO val set.}
    {\begin{tabular}{l|cc|cc|cc|cc|cc|cc}
    \hline
    Backbone & \multicolumn{2}{c|}{ResNet-50} & \multicolumn{2}{c|}{ResNet-152} & \multicolumn{2}{c|}{ViTPose-S} & \multicolumn{2}{c|}{ViTPose-B} & \multicolumn{2}{c|}{ViTPose-L} & \multicolumn{2}{c}{ViTPose-H} \\
    \hline
     Decoder  & Classic & Simple & Classic & Simple & Classic & Simple & Classic & Simple & Classic & Simple & Classic & Simple \\
    \hline
    $AP$    & 71.8  & 53.1  & 73.5  & 55.3  & 73.8  & 73.5 & 75.8  & 75.5  & 78.3  & 78.2  & 79.1  & 78.9  \\
    $AP_{50}$ & 89.8  & 86.9  & 90.5  & 87.9  & 90.1  & 90.0  & 90.7  & 90.6  & 91.4  & 91.4  & 91.7  & 91.6  \\
    \hline
    $AR$    & 77.3  & 62.0  & 79.0  & 63.8  & 79.2  & 78.9  & 81.1  & 80.9  & 83.5  & 83.4  & 84.1  & 84.0  \\
    $AR_{50}$ & 93.7  & 92.1  & 94.3  & 92.9  & 94.0  & 94.0  & 94.6  & 94.6  & 95.3  & 95.3  & 95.4  & 95.4  \\
    \hline
    \end{tabular}}%
  \label{tab:structureSimplicity}%
\end{table*}%

\begin{table}[htbp]
  \centering
  \footnotesize
  \caption{{Ablation study of the decoder in ViTPose on the MS COCO val set. Classic-FP denotes using a randomly initialized and fixed predictor within the Classic decoder (Fig.~\ref{fig:decoderDesign}).}}
    \begin{tabular}{c|cc|cc}
    \hline
          & Classic & Classic-FP & Simple & Minimal \\
    \hline
    $AP$    & 75.8 & 75.6  & 75.5    & 75.4 \\
    $AP_{50}$ & 90.7 & 90.5  & 90.6   & 90.6 \\
    \hline
    $AR$    & 81.1 & 80.9   & 80.9   & 80.7 \\
    $AR_{50}$ & 94.6   & 94.3  & 94.6    & 94.5 \\
    \hline
    \end{tabular}%
  \label{tab:simplerDecoder}%
\end{table}%


\begin{table}[htbp]
  \centering
  \footnotesize
  \caption{The performance of ViTPose-B using different pre-training data on the MS COCO val set .}
    \begin{tabular}{cc|ccc}
    \hline
      Pre-training Dataset & Dataset Volume    & $AP$    & \multicolumn{1}{c}{$AP_{50}$} & \multicolumn{1}{c}{$AP_{75}$}\\
    \hline
    ImageNet-1k  & 1M& 75.8  & 90.7  & 83.2 \\
    COCO (cropping) & 150K & 74.5 & 90.5 & 81.9 \\ 
    COCO+AIC (cropping) & 500K  & 75.8  & 90.8  & 83.0  \\
    COCO+AIC (no cropping) & 300K  & 75.8  & 90.5  & 83.0 \\
    \hline
    \end{tabular}%
  \label{tab:pretrain}%
\end{table}%

\subsection{Implementation details}
\label{sec:implementation}
\textbf{ViTPose.} In the \textbf{top-down paradigm}, ViTPose follows the standard top-down setting for human pose estimation, \ie, a person detector is used to detect person instances and ViTPose is employed to estimate the keypoints for each of the detected instances. The detection results from SimpleBaseline \cite{xiao2018simple} are utilized to evaluate ViTPose's performance on the MS COCO Keypoint val set. We use ViT-S, ViT-B, ViT-L, ViT-H~\cite{dosovitskiy2020image}, and ViTAE-G~\cite{zhang2022vitaev2} as the backbone networks and denote the corresponding models as ViTPose-S, ViTPose-B, ViTPose-L, ViTPose-H, and ViTPose-G, respectively. The backbones are initialized with MAE~\cite{MaskedAutoencoders2021} pre-trained weights, and the models are trained on 8 A100 GPUs based on the MMPose codebase~\cite{mmpose2020}. The default training setting in MMPose is adopted for training the ViTPose models, \ie, we use the $256 \times 192$ input resolution and an AdamW~\cite{reddi2018convergence} optimizer with a learning rate of 5e-4. Udp~\cite{Huang_2020_CVPR} is used for post-processing. The models are trained for 210 epochs, and the learning rate is reduced by multiplying 0.1 at the 170th and 200th epoch, respectively. We sweep each model's layer-wise learning rate decay and stochastic drop path ratio and provide the optimal settings in Table~\ref{tab:implementation}. In the \textbf{Bottom-up paradigm}, we use the whole image as input and resize it to $512 \times 512$ during training, following the common practice in HigherHRNet~\cite{cheng2020higherhrnet}. The models are trained for 300 epochs with an initial learning rate of 1.5e-3. An AdamW~\cite{reddi2018convergence} optimizer is adopted during training. The learning rate is reduced by a factor of 10 at the 200th and 260th epoch, respectively. 

\textbf{ViTPose++.} We also use ViT-S, ViT-B, ViT-L, and ViT-H~\cite{dosovitskiy2020image} as the backbone networks and denote the corresponding models as ViTPose++-S, ViTPose++-B, ViTPose++-L, and ViTPose++-H, respectively. The backbones are initialized with MAE~\cite{MaskedAutoencoders2021} pre-trained weights and the models are trained on 8 A100 GPUs based on the MMPose codebase~\cite{mmpose2020}. The models are trained with the AdamW optimizer~\cite{reddi2018convergence} for 210 epochs. A linear warmup of 500 iterations is also incorporated during training. The learning rate is reduced by multiplying 0.1 at the 170th and 200th epoch, respectively. We randomly sample 1,024 images from all the used training datasets to construct each mini-batch.

\subsection{Ablation studies of ViTPose and analysis}
\label{sec:ablation}
\textbf{The structural simplicity and scalability.} We train ViTPose with the classic decoder and simple decoder as described in Section~\ref{subsec:structuresimplicity}, respectively. We also train SimpleBaseline~\cite{xiao2018simple} with the ResNet~\cite{he2016deep} backbones using the two kinds of decoders for reference. Table~\ref{tab:structureSimplicity} shows the results. It can be observed that using the simple decoder in SimpleBaseline can lead to about 18 AP drops for both ResNet-50 and ResNet-152. However, ViTPose with a plain vision transformer as the backbone and the simple decoder performs well, \ie, only marginal performance drops less than 0.3 AP are observed for either small or large ViTPose models. For the $AP_{50}$ and $AR_{50}$ metrics, ViTPose obtains similar scores regarding both decoders, showing that the plain vision transformer has a strong representation ability to encode the linearly separable features and can thus relieve the necessity of complex decoders. It can also be concluded from the table that the performance of ViTPose improves consistently with the increase of the model size, demonstrating the excellent scalability of ViTPose.

{
\textbf{The influence of simpler decoders.} 
To further investigate the representation capabilities of vision transformer backbones, we incorporate three simplified decoder designs, namely classic-FP, simple, and minimal decoders, into the ViTPose framework. The performance of the ViTPose with these simplified decoder designs is summarized in Table~\ref{tab:simplerDecoder}. Surprisingly, even with the frozen projection layer in the classic-FP or the minimal decoder, ViTPose still achieves an impressive performance of over 75.4 AP. It suggests that a linear layer is enough to act as the task-specific module for pose estimation, demonstrating the excellent representation capability of vision transformers in learning linearly separable features from complex visual data. This emphasizes the notion that \textit{fundamentals speak simply}.
}

\textbf{The influence of pre-training data.} To evaluate whether or not ImageNet data are necessary for pose estimation tasks, we pre-train the transformer backbone on different datasets, \ie, ImageNet-1k~\cite{deng2009imagenet}, MS COCO~\cite{lin2014microsoft}, and a combination of MS COCO and AIC~\cite{wu2017ai}, respectively. Since images in the ImageNet-1k dataset are iconic, we also try to crop the person instances from the MS COCO and AIC training set to form new training data for pre-training. The models are pre-trained for 1,600 epochs on the three datasets, respectively, and then fine-tuned on the MS COCO dataset with pose annotations for 210 epochs. The results are summarized in Table~\ref{tab:pretrain}. It can be seen that with the combination of MS COCO and AIC data for pre-training, ViTPose achieves comparable performance compared with that using ImageNet-1k, while the dataset volume is only half of the ImageNet-1k. It implies that pre-training on the data from downstream tasks has better data efficiency, validating the flexibility of ViTPose in choosing pre-training data. Nevertheless, the AP decreases by 1.3 if only MS COCO data are used for pre-training. It may be caused by the limited volume of the MS COCO dataset, \ie, the number of instances in MS COCO is on 1/3 of the combination of MS COCO and AIC. Besides, there is no obvious benefit of using cropping on the pre-training images by comparing the results in the last two rows, although the larger dataset volume after cropping means a higher training cost. These results further validate that ViTPose has better data efficiency in the pre-training stage when the data come from the same type of tasks as the fine-tuning stage.

\begin{table}[htbp]
  \centering
  \footnotesize
  \caption{{The performance of ViTPose-B using different pre-training methods on the MS COCO val set.}}
    \begin{tabular}{c|cccc}
    \hline
          & Random & DeiT~\cite{touvron2021training}  & MoCov3~\cite{chen2021empirical} & MAE~\cite{MaskedAutoencoders2021} \\
    \hline
    $AP$    & 72.8  & 72.7  & 72.1  & 75.8 \\
    \hline
    \end{tabular}%
  \label{tab:pre-train-method}%
\end{table}%

\begin{table}[htbp]
\footnotesize
  \centering
  \caption{The performance of ViTPose-B regarding different input resolutions on the MS COCO val set.}
    \begin{tabular}{c|cccccc}
    \hline
          & 224x224 & 256x192 & 256x256 & 384x288 & 384x384 & 576x432 \\
    \hline
    $AP$    & 74.9  & 75.8  & 75.8  & 76.9  & 77.1  & 77.8 \\
    $AR$    & 80.4  & 81.1  & 81.1  & 81.9  & 82.0  & 82.6 \\
    \hline
    \end{tabular}%
  \label{tab:inputres}%
\end{table}%

\begin{table*}[htbp]
  \centering
  \footnotesize
  \caption{The performance of ViTPose-B with 1/8 feature size on the MS COCO val set. ``*'' means fp16 is used during training due to the limit of hardware memory. For the combination of vanilla self-attention (Full) and window-based attention (Window), we follow ViTDet~\cite{li2022exploring} and use full attention every 1/4 of all the layers. ``Shift'' and ``Poll'' denote the two strategies described in Section~\ref{subsec:flexibility}.}
    \setlength{\tabcolsep}{0.008\linewidth}{\begin{tabular}{cccc|ccc|cccc}
    \hline
    Full  & Window & Shift & Pool  & Window Size & Training Memory (M) & GFLOPs & $AP$    & $AP_{50}$ & $AR$    & $AR_{50}$ \\
    \hline
    \checkmark     &       &       &       & N/A   & 36,141* & 76.59  & 77.4  & 91.0  & 82.4  & 94.9  \\
    \hline
          & \checkmark     &       &       & (8, 8) & 21,161 & 66.31 & 66.4  & 87.7  & 72.9  & 91.9  \\
          & \checkmark     & \checkmark     &       & (8, 8) & 21,161 & 66.31 & 76.4  & 90.9  & 81.6  & 94.5  \\
          & \checkmark     &       & \checkmark     & (8, 8) & 22,893 & 66.39 & 76.4  & 90.6  & 81.6  & 94.6  \\
          & \checkmark     & \checkmark     & \checkmark     & (8, 8) & 22,893 & 66.39 & 76.8  & 90.8  & 81.9  & 94.8  \\
   \checkmark  & \checkmark     &     &     & (8, 8) & 28,594 & 69.94 & 76.9  & 90.8  & 82.1  & 94.7  \\
    \hline
          & \checkmark     & \checkmark     & \checkmark     & (16, 12) & 26,778 & 68.46 & 77.1  & 91.0  & 82.2  & 94.8  \\
    \hline
    \end{tabular}}%
  \label{tab:attentionmanner}%
\end{table*}%

{

\textbf{The influence of pre-training methods.} We also conduct a comprehensive examination of various pre-training methods to investigate their impact on pose estimation, encompassing random initialization, supervised pre-training, contrastive self-supervised pre-training, and masked image pre-training.
Specifically, we employed DeiT~\cite{touvron2021training} and MoCov3~\cite{chen2021empirical} as the supervised and contrastive self-supervised pre-training methods, respectively, both applied to the ImageNet-1K dataset. The results in Table~\ref{tab:pre-train-method} show that random initialization, supervised pre-training, and contrastive self-supervised pre-training achieve comparable performance on the MS COCO dataset, with random initialization yielding slightly superior performance. This suggests that vision transformers pre-trained using supervised or contrastive self-supervised methods may learn classification-related features that generalize poorly to the pose estimation task.
In contrast, using masked image pre-training, \eg, MAE~\cite{MaskedAutoencoders2021}, significantly enhances the performance.}

\textbf{The influence of input resolution.} To evaluate whether or not ViTPose can adapt well to different input resolutions, we train ViTPose with different input sizes and summarize the results in Table~\ref{tab:inputres}.
The performance of ViTPose-B improves with the increase in the input size. It is also noted that the squared input only brings marginal or even no gains over the rectangular one, \eg, $256 \times 256$ v.s. $256 \times 192$. The reason may be that the average aspect ratio of human instances in MS COCO is about 4:3, and the squared input size does not fit the statistics well.

\textbf{The influence of attention type.} As demonstrated in HRNet~\cite{sun2019deep} and HRFormer~\cite{YuanFHLZCW21}, high-resolution feature maps are beneficial for pose estimation. ViTPose can easily generate high-resolution features by varying the down-sampling ratio of the patching embedding layer, \ie, from 1/16 to 1/8. Besides, to alleviate the out-of-memory issue caused by the quadratic computational complexity of the vanilla self-attention, window-based attention with the shifted window and pooling window strategies described in Section~\ref{subsec:flexibility} is used. The results are presented in Table~\ref{tab:attentionmanner}. Directly using full attention with 1/8 feature size obtains the best 77.4 AP on the MS COCO val set while suffering from a large memory footprint even under the mixed-precision training mode. Window-based attention can alleviate the memory issue while at the cost of performance drop due to lacking global context modeling, \eg, from 77.4 AP to 66.4 AP. The shifted window and pooling window strategies both promote cross-window information exchange for global context modeling and thus significantly improve the performance by 10 AP with less than 10\% memory increase. When applying the two mechanisms together, \ie, the 5th row, the performance further increases to 76.8 AP, which is comparable to the strategy proposed in ViTDet~\cite{li2022exploring} that jointly uses full and window attention (the 6th row), while having less memory footprint, \ie, 76.8 AP v.s. 76.9 AP and 22.9G memory v.s. 28.6G memory. Comparing the 5th and last row in Table~\ref{tab:attentionmanner}, we also note that the performance can be further improved from 76.8 AP to 77.1 AP by enlarging the window size from $8 \times 8$ to $16 \times 12$, which outperforms the ViTDet setting and is comparable with the full attention setting in the first row while having fewer computations and less memory footprint.

\begin{table}[htbp]
  \centering
  \footnotesize
  \caption{The performance of ViTPose-B at three partially fine-tuning settings on the MS COCO val set.}
    \setlength{\tabcolsep}{0.008\linewidth}{\begin{tabular}{cc|cc|cccc}
    \hline
    FFN  & MHSA   &  Memory (M) & GFLOPs & $AP$    & $AP_{50}$ & $AR$    & $AR_{50}$  \\
    \hline
    \checkmark     & \checkmark     & 14,090 & 17.1 & 75.8  & 90.7  & 81.1  & 94.6 \\
    \checkmark     &       & 11,052 & 10.9 & 75.1  & 90.5  & 80.3  & 94.4 \\
          & \checkmark     & 10,941 & 6.2 & 72.8  & 89.8  & 78.3  & 93.8 \\
    \hline
    \end{tabular}}%
  \label{tab:partialtrain}%
\end{table}%

\textbf{The influence of partially fine-tuning.}\label{sec:partial} To assess whether or not vision transformers can still perform well for pose estimation via partially fine-tuning, we fine-tune the ViTPose-B model at three settings, \ie, fully fine-tuning, freezing the MHSA and freezing the FFN. As shown in Table~\ref{tab:partialtrain}, with the MHSA frozen, the performance drops moderately compared with the fully fine-tuning setting, \ie, 75.1 AP v.s. 75.8 AP. The $AP_{50}$ metric is almost the same for the two settings. However, there is a significant drop of 3.0 AP when freezing the FFN. This finding implies that the FFN of transformers is more responsible for task-specific modeling. In contrast, the MHSA is insensitive to different tasks, \eg, probably being only responsible for modeling the relationship of tokens based on feature similarity no matter in the MIM pre-training task or the down-stream pose estimation task.

\begin{table}[htbp]
  \centering
  \footnotesize
  \caption{The performance of knowledge distillation from ViTPose-L to ViTPose-B on the MS COCO val set.}
    \setlength{\tabcolsep}{0.01\linewidth}{\begin{tabular}{cc|c|cccc}
    \hline
    Heatmap & Token  & Memory (M) & $AP$    & $AP_{50}$ & $AR$    & $AR_{50}$ \\
    \hline
    - & -  & 14,090 & 75.8  & 90.7  & 81.1  & 94.6  \\
    \hline
              & \checkmark      & 14,203  & 76.0  & 90.7  & 81.3  & 94.8  \\
    \checkmark     &       & 15,458 & 76.3  & 90.8  & 81.5  & 94.8  \\
    \checkmark     & \checkmark  & 15,565  & 76.6  & 90.9  & 81.8  & 94.9  \\
    \hline
    \end{tabular}}%
  \label{tab:transfer}%
\end{table}%

\textbf{The analysis of transferability.} To evaluate the transferability of ViTPose, we use both the classic output distillation method and the proposed knowledge token distillation method to transfer the knowledge from ViTPose-L to ViTPose-B. The results are listed in Table~\ref{tab:transfer}. As can be seen, the token-based distillation brings a gain of 0.2 AP with a marginal cost of extra memory footprint. In comparison, the output distillation brings a gain of 0.5 AP with a moderate cost of extra memory footprint. It is noteworthy that the proposed token-based distillation does not require the teacher model to be served during the training of the student model, thereby bringing fewer computations and less memory footprint compared with the output distillation method. The two distillation methods are complementary, and using them together obtains 76.6 AP, validating the excellent transferability of ViTPose models.

\subsection{Ablation studies of ViTPose++ and analysis}
\subsubsection{Different settings of ViTPose++}
\label{subsubsec:ViTPose++settings}
\textbf{The baseline method.} Since the decoder in ViTPose is rather simple and lightweight, we can easily extend ViTPose to deal with multiple types of body pose estimation tasks by using a shared backbone and individual decoder for each task. {Specifically, images from each pose estimation task are randomly selected and organized into a batch. These images are then fed into a shared encoder network for feature extraction. The extracted features are subsequently passed to the respective task-specific decoder, which is responsible for predicting the keypoints specific to the given task. This straightforward extension serves as the baseline method and is illustrated in Fig.~\ref{fig:framework} (d).
We gradually introduce various datasets such as MS COCO~\cite{lin2014microsoft}, AIC~\cite{wu2017ai}, MPII~\cite{andriluka20142d}, COCO-W~\cite{jin2020whole}, AP-10K~\cite{yu2021ap10k}, and APT-36K~\cite{yang2022apt} to formulate the training data.
} The results on the MS COCO val set are listed in Table~\ref{tab:multiTask}. Note that we directly use the models after multi-task training for evaluation without further fine-tuning. It can be observed that the performance of ViTPose increases consistently from 75.8 AP to 77.1 AP by using human pose estimation datasets (MS COCO, AIC, and MPII) for training. Although the dataset volume of MPII is much smaller than the combination of MS COCO and AIC (40K v.s. 500K), using MPII for training still brings a 0.1 AP increase, showing that ViTPose can well harness the diverse data in different datasets. With the same data but novel annotations in COCO-W for training, there is no performance gain or drop, showing that ViTPose can well encode the human body feature representations and mitigate the side effect of allocating the representation capacity for those distinct keypoints in the face, hands, and feet. However, with the animal datasets (AP-10K and APT-36K) involved, the performance of the baseline method drops from 77.0 to 76.7, probably due to the conflict between different tasks (\ie, humans and animals).

\begin{table}[htbp]
  \centering
  \footnotesize
  \caption{The performance of ViTPose-B at the multi-task training setting on the MS COCO val set.}
    \setlength{\tabcolsep}{0.008\linewidth}{\begin{tabular}{cccccc|cccc}
    \hline
    COCO  & AIC   & MPII  & COOC-W & AP10K & APT36K & $AP$    & $AP_{50}$  & $AR$    & $AR_{50}$ \\
    \hline
    \checkmark      &  -  &   -   &  -  &  - &   - & 75.8  & 90.7  & 81.1  & 94.6  \\
    \checkmark      & \checkmark &  -   &  -  &  -  & -& 77.0  & 90.8  & 82.2  & 94.9  \\
    \checkmark      & \checkmark &\checkmark & -  &  -  &  - & 77.1  & 90.8  & 82.2  & 94.7  \\
    \checkmark      & \checkmark & \checkmark &\checkmark & -  &  - & 77.1  & 90.8  & 82.2  & 94.8  \\
    \checkmark      & \checkmark & -  & -  & \checkmark & -  & 76.8  & 90.8  & 82.0  & 94.7  \\
    \checkmark      & \checkmark & -  & - & \checkmark & \checkmark & 76.7  &  90.7 &  81.8 & 94.6 \\
    \hline
    \end{tabular}}%
  \label{tab:multiTask}%
\end{table}%

\textbf{ViTPose++.} We consider three variants of ViTPose++ by changing the configurations of the experts in FFN. \textbf{{1) Independent FFN (I-FFN)}}: To mitigate the conflict between different tasks, we use an independent for each task in each transformer block. In this way, each FFN can only process the images of the dataset corresponding to a specific pose estimation task. All these independent FFNs are initialized with the weights of the original FFN from the MAE pre-trained models. \textbf{{2) Independent and Shared FFN (IS-FFN)}}: Although there may exist conflicts between different tasks, they may also share some common knowledge of body poses, \eg, the locations of eyes are symmetric for both human beings and most kinds of animals. To encode such common knowledge, we introduce a shared FFN in each transformer block in addition to the independent ones. In this way, each image will be processed by not only a task-specific FFN but also the shared FFN. The output features from a task-specific FFN and the shared FFN are then summed together and used as the input to the next transformer layer. We initialize the shared FFN with the weights from the MAE pre-trained models and the independent FFNs with zero. \textbf{{3) Partially Shared FFN (PS-FFN)}}: We also explore another design choice to jointly encode the common knowledge and task-specific knowledge of body poses, \ie, the default setting of ViTPose++. As described in Section~\ref{subsec:methodMoE}, we split the last linear layer of each FFN into a shared part and an independent part along the channel dimension and leave the first linear layer shared for all tasks. We duplicate the independent part for all the tasks while not sharing their weights during training. We initialize the shared and independent parts with the weights from the MAE pre-trained models. The features from the first linear layer are processed by the shared part and the corresponding independent part of the second layer, respectively. The output features are then concatenated along the channel dimension and used as the input to the next transformer layer.

\begin{table}[htbp]
  \centering
  \footnotesize
  \caption{The performance of ViTPose++-B at different settings on the MS COCO val set. ViTPose-B is only trained on the MS COCO dataset. ``MT Baseline'' denotes the multi-task training baseline method (Section~\ref{subsubsec:ViTPose++settings}) based on ViTPose-B on the combination of MS COCO, AIC, and AP-10K datasets. The ViTPose++-B at the rest settings are described in Section~\ref{subsubsec:ViTPose++settings} and trained on the same datasets as ``MT Baseline''. The number of parameters during inference is also listed in the second column. 
  }
    \begin{tabular}{c|c|cccc}
    \hline
    Model  & Params (M) & $AP$    & $AP_{50}$  & $AR$    & $AR_{50}$ \\
    \hline
    ViTPose-B & 86  & 75.8  & 90.7  & 81.1  & 94.6 \\
    \hline
    MT Baseline & 86  & 76.8  & 90.8  & 82.0  & 94.7  \\
    \hline
    I-FFN & 86  & 75.8  & 90.5  & 81.1  & 94.4 \\
    IS-FFN & 143 & 77.0  & 90.8  & 82.2  & 94.8 \\
    PS-FFN ($\alpha=1/6$) & 86  & 77.0  & 90.8  & 82.0    & 94.7 \\
    \textbf{PS-FFN ($\alpha=1/4$)} & \textbf{86}  & \textbf{77.0}  & \textbf{90.9}  & \textbf{82.2}  & \textbf{94.8} \\
    PS-FFN ($\alpha=1/3$) & 86  & 77.0  & 90.8  & 82.1  & 94.7 \\
    \hline
    \end{tabular}%
  \label{tab:MoE_FFN}%
\end{table}%

We use the combination of MS COCO, AIC, and AP-10K datasets to train the multi-task baseline model and ViTPose++ at the above three settings. The results are summarized in Table~\ref{tab:MoE_FFN}. It can be observed that without modeling the common knowledge between different tasks, the performance of ViTPose++ with I-FFN obtains 75.8 AP, which is only comparable to the results of ViTPose trained on MS COCO and much worse than the results of the multi-task baseline method (76.8 AP). Although the MHSA layers are shared by all datasets in ViTPose++ with I-FFN, they are almost task-agnostic, as also evidenced in Table~\ref{tab:partialtrain}, thus explaining the unsatisfactory result of I-FFN. {With PS-FFN, ViTPose++ obtains 77.0 AP, which is the same as the performance of using human datasets only for training (2nd row in Table~\ref{tab:multiTask}). The results demonstrate that a proper design for encoding the common and task-specific knowledge of the body poses matters for addressing the conflict between different tasks.} Nevertheless, it introduces extra computations and parameters compared with single-task ViTPose and the multi-task baseline since there are additional FFNs used for each task. With a proper partition ratio $\alpha$ in the PS-FFN setting, \ie, 0.25, ViTPose++ with PS-FFN achieves a better trade-off between performance and model complexity, \ie, obtaining 77.0 AP without extra computations and parameters. Considering that ViTPose++ can be easily extended to handle more types of pose estimation tasks, the results show its flexibility and potential for building a foundation model towards generic body pose estimation.

\subsection{Comparison with SOTA methods}
\subsubsection{The performance on MS COCO}
\begin{table*}[htbp]
  \centering
  \footnotesize
  \caption{Comparison of ViTPose and ViTPose++ with SOTA methods on the MS COCO val set. 
  }
    \setlength{\tabcolsep}{0.007\linewidth}{\begin{tabular}{c|c|ccccccc|cccc}
    \hline
    \multirow{2}[2]{*}{Model} & \multirow{2}[2]{*}{Backbone} & Params & Speed & {FLOPs} & {Memory} & Input & Feature & Detector & \multicolumn{4}{c}{COCO val}  \\
          &       &   (M)    &   (fps)    & {(G)} & {(M)} & Resolution & Resolution & & $AP$  & {$AP_M$} & {$AP_L$} & $AR$    \\
    \hline
    SimpleBaseline~\cite{xiao2018simple} & ResNet-152 & 60  & 829 &  15.7 & 3827  & 256x192 & 1/32 & Faster RCNN & 73.5 & 69.9 & 80.2 & 79.0   \\
    HRNet~\cite{sun2019deep} & HRNet-W32 & 29    & 428 & 16.0 & 7049 & 384x288 & 1/4 & Faster RCNN & 75.8 & 71.9 & 82.8 & 81.0  \\
    HRNet~\cite{sun2019deep} & HRNet-W48 & 64    & 309 & 32.9 & 7339 & 384x288 & 1/4 & Faster RCNN & 76.3 & 72.3 & 83.4 & 81.2  \\
    UDP~\cite{Huang_2020_CVPR}   & HRNet-W48 & 64    &  309  & 32.9 & 7339 & 384x288 & 1/4 & Faster RCNN & 77.2 & 73.2 & 84.4 &  82.0  \\
    {FastPose~\cite{alphapose}} & {ResNet-152} & {60} & {653} &  {16.0} &  {6013} & {256x192} & {1/32} & {YoLov3} & {73.3} & - & - & {-} \\
    TokenPose-L/D24~\cite{li2021tokenpose} & HRNet-W48 & 28    & 602  & 11.0 & 3477 & 256x192 & 1/4 & Faster RCNN & 75.8 & 72.3 & 82.7 & 80.9 \\
    TransPose-H/A6~\cite{yang2021transpose} & HRNet-W48 & 18    & 309 & 21.8 & - & 256x192 & 1/4 & Faster RCNN & 75.8 & 76.4 & 87.2 & 80.8 \\
    HRFormer-B~\cite{YuanFHLZCW21} & HRFormer-B & 43    & 158  & 12.2 &  4287 & 256x192 & 1/4 & Faster RCNN & 75.6 & 71.7 & 82.6 & 80.8 \\
    HRFormer-B~\cite{YuanFHLZCW21} & HRFormer-B & 43    & 78 & 26.8 & 7859 & 384x288 & 1/4 & Faster RCNN & 77.2 & 73.2 & 84.2 & 82.0  \\
    \hline
    ViTPose-S & ViT-S & 22    & 1439& 5.3 & 3438 & 256x192 & 1/16 & Faster RCNN & 73.8 & 70.5 & 80.4 & 79.2  \\
    ViTPose-B & ViT-B & 86    & 944  & 17.1 & 4589  & 256x192 & 1/16 & Faster RCNN & 75.8 & 72.1 & 82.2 & 81.1 \\
    ViTPose-L & ViT-L & 307 & 411 & 59.8 & 5587 & 256x192 & 1/16 & Faster RCNN & 78.3 & 74.5 & 85.4 & 83.5 \\
    ViTPose-H & ViT-H & 632 & 241 & 122.9 & 7293 & 256x192 & 1/16 & Faster RCNN & 79.1 & 75.3 & 86.0 & 84.1  \\
    \hline
    ViTPose++-S & ViT-S & 22    & 1439 & 5.3 & 3438 & 256x192 & 1/16 & Faster RCNN & 75.8 & 72.3 & 82.6 & 81.0  \\
    ViTPose++-B & ViT-B & 86    & 944 & 17.1 & 4589 & 256x192 & 1/16 & Faster RCNN & 77.0 & 73.4 & 84.0 & 82.6  \\
    ViTPose++-L & ViT-L & 307 & 411 & 59.8 & 5587 & 256x192 & 1/16 & Faster RCNN & 78.6 & 75.2 & 85.6 & 84.1 \\
    ViTPose++-H & ViT-H & 632 & 241 & 122.9 & 7293 & 256x192 & 1/16 & Faster RCNN & 79.4 & 75.8 & 86.5 & 84.8  \\
    \hline
    \end{tabular}}%
  \label{tab:ComparisonValTest}%
\end{table*}%

Since MS COCO~\cite{lin2014microsoft} is the most popular and representative dataset among all the body pose estimation tasks, we first compare the performance of ViTPose and ViTPose++ with SOTA methods on the MS COCO dataset. Specifically, we report their results at both the top-down and bottom-up paradigms. Then, we further compare their performance on other body pose estimation datasets. 

\textbf{Top-down paradigm.} Based on the previous analysis, we use the input resolution of $256 \times 192$ during training and report the results on the MS COCO val set as shown in Table~\ref{tab:ComparisonValTest}. The speed of all methods is recorded on a single A100 GPU with a batch size of 64. The ViTPose++ models are trained on the combination of MS COCO, AIC, MPII, COCO-W, AP-10K, and APT-36K datasets. It can be observed that ViTPose achieves a better trade-off between throughput and accuracy, showing that the plain vision transformer has a strong representation ability and is computationally friendly to modern hardware devices. {For example, the ViTPose-S model with fewer parameters and fewer memory footprint obtains similar performance compared to SimpleBaseline~\cite{xiao2018simple} and FastPose~\cite{alphapose} using a ResNet-152~\cite{he2016deep} backbone. Besides, ViTPose performs better with much larger backbones, demonstrating the good scalability of ViTPose.} For example, ViTPose-L obtains much better performance than ViTPose-S and ViTPose-B, \ie, 78.3 AP v.s. 73.8 AP and 75.8 AP. ViTPose-L also outperforms previous SOTA methods based on CNN and transformers (\eg, UDP~\cite{Huang_2020_CVPR} and TokenPose~\cite{li2021tokenpose}) by a large margin while keeping a similar inference speed. {Furthermore, in comparison to the strong transformer-based baseline method HRFormer-B~\cite{YuanFHLZCW21}, the proposed plain vision transformer-based ViTPose delivers comparable performance. For instance, ViTPose-H (16th row) achieves slightly better performance and faster inference speed compared to HRFormer-B (10th row), with 79.1 AP v.s. 75.6 AP and 241 fps v.s. 158 fps, respectively. Even when compared to HRFormer-B with a larger input resolution (11th row), ViTPose-H still outperforms it while consuming less memory. These results highlight that despite having a higher number of parameters, the plain vision transformer structure, with its simple matrix multiplication operations, exhibits excellent compatibility with modern hardware designs. It also has the potential to achieve better performance with reduced memory consumption and faster speed, pointing towards a new design paradigm for future works.} While using multiple body pose estimation datasets for training, the performance of ViTPose++ further increases. For example, the small model ViTPose++-S obtains 75.8 AP with 1,439 fps, which is comparable to the larger model ViTPose-B but has a much faster speed. \rev{It's also worth noting that even when applied to large-scale models that have already achieved outstanding results, ViTPose++-H still demonstrates performance improvement by elevating ViTPose-H from an impressive 79.1 AP to an even higher 79.4 AP.} The results show the good scalability and flexibility of ViTPose regarding both model structures and training data. 

\begin{table}[htbp]
\footnotesize
  \centering
  \caption{Comparison with SOTA methods on the MS COCO test-dev set. ``$\diamond$'' means model ensemble. ``$\dagger$'', ``$\ddagger$'', and ``*'' denote the champions of the 2018, 2019, and 2020 MS COCO Human Keypoint Detection Challenge, respectively.}
    \setlength{\tabcolsep}{0.01\linewidth}{\begin{tabular}{llcccccc}
    \hline
      Method  & Backbone & $AP$    & $AP_{50}$  & $AP_{75}$  & $AP_M$   & $AP_L$   & $AR$ \\
    \hline
    Baseline$^\diamond$ \cite{xiao2018simple} & ResNet-152 &76.5     &  92.4     &  84.0     &  73.0     & 82.7     & 81.5 \\
    {RMPE~\cite{fang2017rmpe}} & {PyraNet} & {68.8} & {87.5} & {75.9} & {64.6} & {75.1} & {73.6} \\
    {CPN+~\cite{chen2018cascaded}} & {ResNet-Inception} & {73.0} & {91.7} & {80.9} & {69.5} & {78.1} & {79.0} \\
    HRNet \cite{sun2019deep} & HRNet-w48 &  77.0   &  92.7     &  84.5     &  73.4     &  83.1    & 82.0 \\
    {HRFormer~\cite{YuanFHLZCW21}} & {HRFormer-B} & {76.2} &  {92.7} & {83.8} & {72.5} & {82.3} & {81.2} \\
    MSPN$^{\diamond\dagger}$ \cite{li2019rethinking} & 4xResNet-50 &78.1  &{94.1}       &{85.9}       &74.5       &{83.3}       &{83.1} \\
    SwinV2~\cite{xie2022revealing} & SwinV2-L & 77.2 & - & - & - & - & - \\
    DARK \cite{zhang2020distribution} & HRNet-w48 & 77.4 &  92.6     &  84.6     &  73.6     &  83.7    & 82.3 \\
    RSN$^{\diamond\ddagger}$ \cite{cai2020learning} & 4xRSN-50 &{79.2}  &{94.4}       &{87.1}       &{76.1}       &{83.8}       &{84.1} \\
    CCM$^\diamond$~\cite{zhang2021towards} & HRNet-w48& {78.9}    &  {93.8}     &  {86.0}     &  {75.0}     &  {84.5}    & {83.6} \\
    UDP++$^{\diamond*}$~\cite{Huang_2020_CVPR} & HRNet-w48plus & 80.8 & 94.9 & 88.1 & 77.4 & 85.7	& 85.3 \\
    \hline
    ViTPose-B & ViT-B & 75.1 & 92.5 & 83.1 & 72.0 & 80.7 & 80.3 \\
    ViTPose-L & ViT-L & 77.3 & 93.1 & 85.3 & 74.0 & 83.1 & 82.4 \\
    ViTPose-H & ViT-H & 78.1 & 93.3 & 85.7 & 74.9 & 83.8 & 83.1 \\
    \hline
    \rev{ViTPose++-B}  & \rev{ViT-B}  & \rev{76.4}  & \rev{92.7}  & \rev{84.3}  & \rev{73.2}  & \rev{82.2} & \rev{81.5} \\
    \rev{ViTPose++-L}  & \rev{ViT-L}  & \rev{77.8}  & \rev{93.1}  & \rev{85.5}  & \rev{74.6}  & \rev{83.6} & \rev{82.9} \\
    \rev{ViTPose++-H}  & \rev{ViT-H} & \rev{78.5}  & \rev{93.4}  & \rev{86.2}  & \rev{75.3}  & \rev{84.4} & \rev{83.4} \\
    \hline
    ViTPose-G & ViTAE-G & 80.9 & 94.8 & 88.1 & 77.5 & 85.9 & 85.4 \\
    {ViTPose-G$^{\diamond}$} & {ViTAE-G} & {81.1} & {95.0} &	{88.2} & {77.8} & {86.0} & {85.6}\\
    \hline
    \end{tabular}}%
  \label{tab:SOTA_test}%
\end{table}%

We then build a much stronger ViTPose-G model using the ViTAE-G~\cite{zhang2022vitaev2} backbone, which has about 1B parameters and larger input resolution ($576 \times 432$) and trained on the combination of MS COCO, AIC, and MPII datasets. A more powerful detector from Bigdet~\cite{bigdetection2022} is also used to provide person detection results (68.5 AP on the person class of COCO dataset). As shown in Table~\ref{tab:SOTA_test}, ViTPose-G with the ViTAE-G backbone outperforms all previous SOTA methods on the MS COCO test-dev set at 80.9 AP, where the previous best method UDP++ uses 17 models and a slightly better detector (68.6 AP on the person class of COCO dataset) and only obtains 80.8 AP. Our ensemble model ViTPose-G$^{\diamond}$ with only three models further achieves the best 81.1 AP.

\begin{table}[htbp]
  \centering
  \footnotesize
  \caption{Comparison of ViTPose and SOTA methods in the bottom-up paradigm on the MS COCO val set.}
    \setlength{\tabcolsep}{0.015\linewidth}{\begin{tabular}{c|c|cccc}
    \hline
    \multirow{2}[2]{*}{Model} & \multirow{2}[2]{*}{Backbone} & \multicolumn{4}{c}{COCO Val} \\
          &       & $AP$    & $AP_{50}$  & $AR$    & $AR_{50}$ \\
    \hline
    Associate Embedding~\cite{newell2017associative} & Hourglass  & 61.3  & 83.3  & 65.9  & 85.0 \\
    Associate Embedding~\cite{newell2017associative} & ResNet-50   & 46.6  & 74.2  & 56.6  & 81.0 \\
    Associate Embedding~\cite{newell2017associative} & ResNet-101  & 55.4  & 80.7  & 62.2  & 84.1 \\
    Associate Embedding~\cite{newell2017associative} & ResNet-152  & 59.5  & 82.9  & 65.1  & 85.6 \\
    {Pifpaf~\cite{kreiss2019pifpaf}} & {ResNet-50} & {62.6} & - & - & - \\
    HigherHRNet-w32~\cite{cheng2020higherhrnet} & HRNet-w32  & 67.7  & 87.0  & 72.3  & 89.0 \\
    HigherHRNet-w48~\cite{cheng2020higherhrnet} & HRNet-w48   & 68.6  & 87.3  & 73.1  & 89.2 \\
    \hline
    ViTPose-B & ViT-B   & 68.5  & 87.3  & 72.9  & 89.2 \\
    ViTPose-L & ViT-L   & 70.1  & 88.3  & 74.5  & 90.2 \\
    \hline
    \end{tabular}}%
  \label{tab:BottomUpResults}%
\end{table}%

\textbf{Bottom-up paradigm.} We use images of size $512 \times 512$ in MS COCO to train the ViTPose models in the bottom-up paradigm and report the performance on the MS COCO val set. 1/8 feature resolutions with full window-based attention are adopted. The results are listed in Table~\ref{tab:BottomUpResults}. Without delicate designs, the plain vision transformer obtains 68.5 AP with ViTPose-B and 70.1 AP with ViTPose-L, demonstrating the good scalability of ViTPose in the bottom-up paradigm. Compared with previous SOTA methods, \eg, HigherHRNet with the HRNet-w48 backbone, ViTPose-L model achieves better performance and sets a new SOTA, \ie, 68.6 AP v.s. 70.1 AP. These results validate the flexibility of plain vision transformers for different keypoint detection paradigms.

\begin{table*}[htbp]
  \centering
  \footnotesize
  \caption{Comparison of ViTPose++ and SOTA methods on the OCHuman~\cite{zhang2019pose2seg} val and test set with ground truth bounding boxes.}
    {\begin{tabular}{c|cc|cccc|cccc}
    \hline
    \multirow{2}[4]{*}{Model} & \multirow{2}[4]{*}{Backbone} & \multirow{2}[4]{*}{Resolution} & \multicolumn{4}{c|}{Val Set}  & \multicolumn{4}{c}{Test Set} \\
\cline{4-11}          &       &       & $AP$    & $AP_{50}$  & $AR$    & $AR_{50}$  & $AP$    & $AP_{50}$  & $AR$    & $AR_{50}$ \\
    \hline
    SimpleBaseline~\cite{xiao2018simple} & ResNet-152 & 384x288 & 58.8  & 72.7  & 63.1  & 75.7  & 58.2  & 72.3  & 62.7  & 75.2  \\
    HRNet~\cite{sun2019deep} & HRNet-w32 & 384x288 & 60.9  & 76.0  & 65.1  & 78.2  & 60.6  & 74.8  & 64.7  & 77.6  \\
    HRNet~\cite{sun2019deep} & HRNet-w48 & 384x288 & 62.1  & 76.1  & 65.9  & 78.2  & 61.6  & 74.9  & 65.3  & 77.3  \\
    MIPNet~\cite{mipnet} & HRNet-w48 & 384x288 & 74.1  & 89.7  & 81.0  & -     & -     & -     & -     & - \\
    HRFormer~\cite{YuanFHLZCW21} & HRFormer-S & 384x288 & 53.1  & 73.1  & 59.6  & 76.9  & 52.8  & 72.8  & 59.1  & 76.6  \\
    HRFormer~\cite{YuanFHLZCW21} & HRFormer-B & 384x288 & 50.4  & 71.5  & 58.8  & 76.6  & 49.7  & 71.6  & 58.2  & 76.0  \\
    \hline
    \rev{ViTPose-S} & \rev{ViT-S} & \rev{256x192} & \rev{57.6}  & \rev{75.2}  & \rev{61.8}  & \rev{77.8}  & \rev{57.4}  & \rev{74.7}  & \rev{61.8}  & \rev{77.4}  \\
    \rev{ViTPose-B} & \rev{ViT-B} & \rev{256x192} & \rev{60.1} & \rev{75.9} & \rev{64.3} & \rev{78.0} & \rev{59.6} & \rev{74.7} & \rev{64.7} & \rev{77.8}  \\
    \rev{ViTPose-L} & \rev{ViT-L} & \rev{256x192} & \rev{65.2} & \rev{77.2} & \rev{69.1} & \rev{79.8} & \rev{64.2} & \rev{76.0} & \rev{67.8} & \rev{78.4}  \\
    \rev{ViTPose-H} & \rev{ViT-H} & \rev{256x192} & \rev{67.5} & \rev{79.6} & \rev{70.7} & \rev{81.1} & \rev{67.0} & \rev{78.5} & \rev{70.2} & \rev{80.7}  \\
    \hline
    ViTPose++-S & ViT-S & 256x192 & 79.4  & 90.7  & 81.6  & 91.4  & 78.4  & 90.6  & 80.6  & 91.0  \\
    ViTPose++-B & ViT-B & 256x192 & 83.7  & 91.8  & 85.4  & 92.9  & 82.6  & 91.7  & 84.6  & 92.4  \\
    ViTPose++-L & ViT-L & 256x192 & 87.4  & 93.7  & 88.8  & 94.2  & 85.7  & 92.8  & 87.5  & 93.4  \\
    ViTPose++-H & ViT-H & 256x192 & 86.8  & 92.8  & 88.3  & 93.7  & 85.7  & 92.8  & 87.4  & 93.8  \\
    \hline
    \end{tabular}}%
  \label{tab:ochuman}%
\end{table*}%

\begin{table*}[htbp]
  \centering
  \caption{Comparison (PCKh) of ViTPose++ and SOTA methods on the MPII~\cite{andriluka20142d} val set with ground truth bounding boxes.}
  \footnotesize
    \setlength{\tabcolsep}{0.005\linewidth}{\begin{tabular}{c|cc|ccccccc|c}
    \hline
    Model & Backbone & Resolution & Head  & Shoulder & Elbow & Wrist & Hip   & Knee  & Ankle & Mean \\
    \hline
    SimpleBaseline~\cite{xiao2018simple} & ResNet-152 & 256x256 & 86.9  & 95.4  & 89.4  & 84.0  & 88.0  & 84.6  & 82.1  & 89.0 \\
    HRNet~\cite{sun2019deep} & HRNet-w32 & 256x256 & 96.9  & 85.9  & 90.5  & 85.9  & 89.1  & 86.1  & 82.5  & 90.0  \\
    HRNet~\cite{sun2019deep} & HRNet-w48 & 256x256 & 97.1  & 95.8  & 90.7  & 85.6  & 89.0  & 86.8  & 82.1  & 90.1   \\
    CFA~\cite{CFA}   & ResNet-101 & 384x384 & 95.9  & 95.4  & 91.0  & 86.9  & 89.8  & 87.6  & 83.9  & 90.1 \\
    ASDA~\cite{ASDA}  & HRNet-w48 & 256x256 & 97.3  & 96.5  & 91.7  & 87.9  & 90.8  & 88.2  & 84.2  & 91.4  \\
    TransPose-H-A6~\cite{yang2021transpose} & HRNet-w48 & 256x256 & -     & -     & -     & -     & -     & -     & -     & 92.3  \\
    {OKDHP~\cite{li2021online}} & {8-Stack Hourglass} & {256x256} & {97.3} & {96.1} & {91.2} & {86.8} & {89.9} & {86.9} & {83.1} & {90.6} \\
    {HRFormer~\cite{YuanFHLZCW21}} & {HRFormer-S} & {256x256} & {97.1}    & {95.8}     & {90.5}     & {85.9}     & {88.7}     & {85.7}  &  {82.1}  & {89.9} \\
    {HRFormer~\cite{YuanFHLZCW21}} & {HRFormer-B} & {256x256} & {96.8}     & {96.1}     & {90.4}     & {85.9}     & {89.0}     & {87.3}     & {84.1}     & {90.4} \\
    \hline
    \rev{ViTPose-S} & \rev{ViT-S} & \rev{256x192} & \rev{96.4} & \rev{94.7} & \rev{88.1} & \rev{83.2} & \rev{88.4} & \rev{84.3} & \rev{80.0} & \rev{88.4}   \\
    \rev{ViTPose-B} & \rev{ViT-B} & \rev{256x192} & \rev{97.0} & \rev{96.2} & \rev{90.7} & \rev{86.7} & \rev{90.4} & \rev{88.2} & \rev{84.2} & \rev{90.9}   \\
    \rev{ViTPose-L} & \rev{ViT-L} & \rev{256x192} & \rev{97.7} & \rev{97.2} & \rev{92.9} & \rev{89.2} & \rev{92.3} & \rev{90.8} & \rev{87.4} & \rev{92.8}  \\
    \rev{ViTPose-H} & \rev{ViT-H} & \rev{256x192} & \rev{97.7} & \rev{97.1} & \rev{93.2} & \rev{89.6} & \rev{91.8} & \rev{91.2} & \rev{88.1} & \rev{93.0}  \\
    \hline
    ViTPose++-S & ViT-S & 256x192 & 97.4  & 97.2  & 92.9  & 89.0  & 92.3  & 90.4  & 86.8 & 92.7   \\
    ViTPose++-B & ViT-B & 256x192 & 97.3  & 97.2  & 93.3  & 89.7  & 91.5  & 90.7  & 87.2 & 92.8   \\
    ViTPose++-L & ViT-L & 256x192 & 98.0  & 97.6  & 94.3  & 90.9  & 92.9  & 92.6  & 89.5  & 94.0  \\
    ViTPose++-H & ViT-H & 256x192 & 97.8  & 97.6  & 94.4  & 91.5  & 93.2  & 92.8  & 90.2 & 94.2  \\
    \hline
    \end{tabular}}%
  \label{tab:mpii}%
\end{table*}%

\begin{table*}[htbp]
  \centering
  \caption{Comparison of ViTPose++ and SOTA methods on the AIC~\cite{wu2017ai} val set with ground truth bounding boxes.}
  \footnotesize
    \begin{tabular}{c|cc|ccccc}
    \hline
    Method & Backbone & Resolution & $AP$    & $AP_{50}$ & $AP_{75}$ & $AR$    & $AR_{50}$ \\
    \hline
    SimpleBaseline~\cite{xiao2018simple} & ResNet-50 & 256x192 & 28.0  & 71.6  & 15.8  & 32.1  & 74.1  \\
    SimpleBaseline~\cite{xiao2018simple} & ResNet-101 & 256x192 & 29.4  & 73.6  & 17.4  & 33.7  & 76.3  \\
    SimpleBaseline~\cite{xiao2018simple} & ResNet-152 & 256x192 & 29.9  & 73.8  & 18.3  & 34.3  & 76.9  \\
    HRNet~\cite{sun2019deep} & HRNet-w32 & 256x192 & 32.3  & 76.2  & 21.9  & 36.6  & 78.9  \\
    HRNet~\cite{sun2019deep} & HRNet-w48 & 256x192 & 33.5  & 78.0  & 23.6  & 37.9  & 80.0  \\
    HRFormer~\cite{YuanFHLZCW21} & HRFomer-S & 256x192 & 31.6  & 75.9  & 20.9  & 35.8  & 78.0  \\
    HRFormer~\cite{YuanFHLZCW21} & HRFomer-B & 256x192 & 34.4  & 78.3  & 24.8  & 38.7  & 80.9  \\
    \hline
    \rev{ViTPose-S} & \rev{ViT-S} & \rev{256x192} & \rev{28.2} & \rev{72.4} & \rev{15.8} & \rev{32.4} & \rev{75.1}  \\
    \rev{ViTPose-B} & \rev{ViT-B} & \rev{256x192} & \rev{30.9} & \rev{75.8} & \rev{19.4} & \rev{35.3} & \rev{78.3}  \\
    \rev{ViTPose-L} & \rev{ViT-L} & \rev{256x192} & \rev{34.1} & \rev{79.1} & \rev{23.9} & \rev{38.7} & \rev{81.6}  \\
    \rev{ViTPose-H} & \rev{ViT-H} & \rev{256x192} & \rev{34.6} & \rev{80.2} & \rev{24.3} & \rev{39.0} & \rev{82.1}  \\
    \hline
    ViTPose++-S & ViT-S & 256x192 & 29.7  & 74.6  & 17.6 & 34.3  & 77.1  \\
    ViTPose++-B & ViT-B & 256x192 & 31.8  & 76.7  & 20.6 & 36.3  & 79.0  \\
    ViTPose++-L & ViT-L & 256x192 & 34.3  & 79.1  & 24.1 & 38.9  & 81.8  \\
    ViTPose++-H & ViT-H & 256x192 & 34.8  & 80.2  & 24.5 & 39.1  & 82.2  \\
    \hline
    \end{tabular}%
  \label{tab:aic}%
\end{table*}%

\subsubsection{The performance on other datasets}
To evaluate the performance of ViTPose comprehensively, apart from the results on the MS COCO val and test-dev set, we also report the performance of ViTPose++-S, ViTPose++-B, ViTPose++-L, and ViTPose++-H on the OCHuman~\cite{zhang2019pose2seg} val and test set, MPII~\cite{andriluka20142d} val set, AIC~\cite{wu2017ai} val set, COCO-W~\cite{jin2020whole} val set, AP-10K~\cite{yu2021ap} test set, and APT-36K~\cite{yang2022apt} test set, respectively. Please note that the ViTPose++ models are trained with the combination of all the datasets and directly tested on the target dataset without further fine-tuning, which keeps the whole pipeline as simple as possible. For each dataset, we use the corresponding FFN, decoder, and prediction head in ViTPose for prediction. \rev{We also provide the ViTPose baseline results. It's worth highlighting that, despite using the same number of parameters for inference, ViTPose++ utilizes much fewer parameters during training compared with training individual ViTPose models for each dataset.}

\textbf{OCHuman val and test set.} To evaluate the performance of different methods on the human instances with heavy occlusions, we compare ViTPose++ and SOTA methods on the OCHuman val and test set. We adopt the ground truth bounding boxes instead of those obtained by a person detector to isolate the effect of person detection because not all human instances are annotated in the OCHuman datasets, and the person detector may cause false positive or missing bounding boxes, which may obscure the true performance of pose estimation models. Specifically, we use the decoder and prediction head of ViTPose++ corresponding to the MS COCO dataset since the keypoint definition is the same in both the MS COCO and OCHuman datasets. The results are listed in Table~\ref{tab:ochuman}. Compared with previous SOTA methods with complex structures, \eg, MIPNet~\cite{mipnet}, ViTPose++ obtains a gain of over 10 AP on the OCHuman val set, although there is no special structural design to deal with occlusions, implying the strong feature representation ability of plain vision transformer. It should also be noted that HRFormer~\cite{YuanFHLZCW21} experiences large performance drops from MS COCO to OCHuman, and its small model even beats the base model, \ie, 53.1 AP v.s 50.4 AP on the OCHuman val set. Such phenomena imply that HRFormer may overfit the MS COCO dataset, especially for larger models. By contrast, ViTPose++ generally delivers performance gains with the increase of model size and significantly pushes forward the frontier of the keypoint detection performance on OCHuman, \ie, 87.4 AP on the val set and 85.7 AP on the test set. 

\textbf{MPII val set.} We evaluate the performance of ViTPose++ and representative models on the MPII val set with the ground truth bounding boxes. Following the default settings of MPII, we use PCKh as the metric for performance evaluation. {It can be observed that the previous representative methods have obtained well performance on the MPII dataset, \ie, the strong transformer-based method HRFormer-B obtains over 90 PCKh. Nevertheless, ViTPose demonstrates the potential to further improve performance due to its remarkable scalability.} As shown in Table~\ref{tab:mpii}, ViTPose++ outperforms previous methods in terms of both single joint evaluation and average evaluation, \eg, ViTPose++-S, ViTPose++-B, ViTPose++-L, and ViTPose++-H achieve 92.7, 92.8, 94.0, and 94.2 average PCKh with a smaller input resolution.

\textbf{AIC val set.} We also evaluate the performance of ViTPose++ on the AIC val set. As listed in Table~\ref{tab:aic}, compared to representative CNN-based and transformer-based models, our ViTPose++ obtains better performance, \ie, 34.8 AP by ViTPose-H v.s. 33.5 AP by HRNet-w48 and 34.4 AP by HRFromer-B. Nevertheless, the score is still not high enough on the AIC set, indicating that more efforts may be needed to improve the performance further.

\begin{table*}[htbp]
  \centering
  \caption{Comparison of ViTPose and SOTA methods on the COCO-W~\cite{jin2020whole} val set with detected boxes.}
  \footnotesize
    \begin{tabular}{c|cc|cccc|c}
    \hline
    Method & Backbone & Resolution & Body    & Foot & Face & Hand & WholeBody \\
    \hline
    SimpleBaseline~\cite{xiao2018simple} & ResNet-50 & 256x192 & 65.2  & 61.4  & 60.8  &  46.0 &  52.0 \\
    SimpleBaseline~\cite{xiao2018simple} & ResNet-101 & 256x192 & 67.0  & 64.0  & 61.1  & 46.3  & 53.3  \\
    HRNet~\cite{sun2019deep} & HRNet-w32 & 256x192 & 70.0  & 56.7  & 63.7  & 47.3  & 55.3  \\
    HRNet~\cite{sun2019deep} & HRNet-w48 & 256x192 & 70.0  & 67.2  & 65.6  & 53.4  & 57.9  \\
    \rev{ZoomNet}  & \rev{HRNet}  & \rev{384x288}  & \rev{74.3}  & \rev{79.8}  & \rev{62.3}  & \rev{40.1}  & \rev{54.1} \\
    \rev{FastPose}  & \rev{ResNet50-dcn} & \rev{256x192}  & \rev{69.3}  & \rev{69.0}  & \rev{75.9}  & \rev{45.3}  & \rev{57.7} \\
    \hline
    \rev{ViTPose-S}  & \rev{ViT-S}  & \rev{256x192}  & \rev{66.1}  & \rev{64.5}  & \rev{59.8}  & \rev{45.9}  & \rev{52.3} \\
    \rev{ViTPose-B}  & \rev{ViT-B}  & \rev{256x192}  & \rev{69.6}  & \rev{70.1}  & \rev{62.1}  & \rev{50.8}  & \rev{56.3} \\
    \rev{ViTPose-L}  & \rev{ViT-L}  & \rev{256x192}  & \rev{72.9}  & \rev{75.2}  & \rev{63.9}  & \rev{54.5}  & \rev{59.6} \\
    \rev{ViTPose-H}  & \rev{ViT-H}  & \rev{256x192}  & \rev{73.5}  & \rev{76.4}  & \rev{63.2}  & \rev{54.9}  & \rev{60.0} \\
    \hline
    ViTPose++-S & ViT-S & 256x192 & 71.6  & 72.1  & 55.9  & 45.3 & 54.4 \\
    ViTPose++-B & ViT-B & 256x192 & 73.3  & 74.2  & 60.1  & 50.2 & 57.4 \\
    ViTPose++-L & ViT-L & 256x192 & 75.3  & 77.1  & 63.0 & 54.2 & 60.6 \\
    ViTPose++-H & ViT-H & 256x192 & 75.9  & 77.9  & 63.3 & 54.7 & 61.2 \\
    \hline
    \end{tabular}%
  \label{tab:wholebody}%
\end{table*}%

\begin{table*}[htbp]
  \centering
  \caption{Comparison of ViTPose and SOTA methods on the AP-10K~\cite{yu2021ap} test set with ground truth bounding boxes.}
  \footnotesize
    \begin{tabular}{c|cc|ccccc}
    \hline
    Method & Backbone & Resolution & $AP$    & $AP_{50}$ & $AP_{75}$ & $AP_{M}$    & $AP_{L}$ \\
    \hline
    SimpleBaseline~\cite{xiao2018simple} & ResNet-50 & 256x256 & 68.1  & 92.3  & 74.0  & 51.0  & 68.8  \\
    SimpleBaseline~\cite{xiao2018simple} & ResNet-101 & 256x256 & 68.1  & 92.2  & 74.2  & 53.4  & 68.8  \\
    HRNet~\cite{sun2019deep} & HRNet-w32 & 256x256 & 72.2  & 93.9  & 78.7  & 55.5  & 73.0  \\
    HRNet~\cite{sun2019deep} & HRNet-w48 & 256x256 & 73.1  & 93.7  & 80.4  & 57.4  & 73.8  \\
    \hline
    \rev{ViTPose-S}  & \rev{ViT-S}  & \rev{256x192}  & \rev{68.7}  & \rev{93.0}  & \rev{75.1}  & \rev{50.7}  & \rev{69.1} \\
    \rev{ViTPose-B}  & \rev{ViT-B}  & \rev{256x192}  & \rev{73.4}  & \rev{95.0}  & \rev{81.9}  & \rev{60.2}  & \rev{73.8} \\
    \rev{ViTPose-L}  & \rev{ViT-L}  & \rev{256x192}  & \rev{80.1}  & \rev{97.5}  & \rev{88.0}  & \rev{62.3}  & \rev{80.3} \\
    \rev{ViTPose-H}  & \rev{ViT-H}  & \rev{256x192}  & \rev{82.0}  & \rev{97.5}  & \rev{89.1}  & \rev{67.3}  & \rev{82.3} \\
    \hline
    ViTPose++-S & ViT-S & 256x192 & 71.4  & 93.3  & 78.4 & 47.6 & 71.8 \\
    ViTPose++-B & ViT-B & 256x192 & 74.5  & 94.9  & 82.2 & 46.8 & 75.0 \\
    ViTPose++-L & ViT-L & 256x192 & 80.4  & 97.6  & 88.5 & 52.7 & 80.8 \\
    ViTPose++-H & ViT-H & 256x192 & 82.4  & 98.2  & 89.5 & 59.1 & 82.8 \\
    \hline
    \end{tabular}%
  \label{tab:ap10k}%
\end{table*}%

\begin{table*}[htbp]
  \centering
  \caption{Comparison of ViTPose and SOTA methods on the APT-36K~\cite{yang2022apt} val set with ground truth bounding boxes.}
  \footnotesize
    \begin{tabular}{c|cc|ccccc}
    \hline
    Method & Backbone & Resolution & $AP$    & $AP_{50}$ & $AP_{75}$ & $AR $    & $AR_{50}$ \\
    \hline
    SimpleBaseline~\cite{xiao2018simple} & ResNet-50 & 256x256 & 69.4  & -  & -  &  - &  - \\
    SimpleBaseline~\cite{xiao2018simple} & ResNet-101 & 256x256 & 69.6  & -  & -  & -  & -  \\
    HRNet~\cite{sun2019deep} & HRNet-w32 & 256x256 & 74.2  & -  & -  & -  & -  \\
    HRNet~\cite{sun2019deep} & HRNet-w48 & 256x256 & 74.1  & -  & -  & -  & -  \\
    HRFormer~\cite{sun2019deep} & HRFormer-S & 256x256 & 71.3  & -  & -  & -  & -  \\
    HRFormer~\cite{sun2019deep} & HRFormer-B & 256x256 & 74.2  & -  & -  & -  & - \\
    \hline
    \rev{ViTPose-S}  & \rev{ViT-S}  & \rev{256x192}  & \rev{70.2}  & \rev{94.8}  & \rev{77.1}  & \rev{73.2}  & \rev{95.4} \\
    \rev{ViTPose-B}  & \rev{ViT-B}  & \rev{256x192}  & \rev{75.5}  & \rev{97.4}  & \rev{85.0}  & \rev{78.4}  & \rev{97.6} \\
    \rev{ViTPose-L}  & \rev{ViT-L}  & \rev{256x192}  & \rev{80.6}  & \rev{98.4}  & \rev{90.4}  & \rev{83.3}  & \rev{98.5} \\
    \rev{ViTPose-H}  & \rev{ViT-H}  & \rev{256x192}  & \rev{82.1}  & \rev{98.4}  & \rev{91.6}  & \rev{84.9}  & \rev{98.6} \\
    \hline
    ViTPose++-S & ViT-S & 256x192 & 74.2  & 94.9  & 82.3 & 77.6 & 95.6 \\
    ViTPose++-B & ViT-B & 256x192 & 75.9  & 95.4  & 83.7 & 79.2 & 96.0 \\
    ViTPose++-L & ViT-L & 256x192 & 80.8  & 97.4  & 88.5 & 83.9 & 97.9 \\
    ViTPose++-H & ViT-H & 256x192 & 82.3  & 97.7  & 90.6 & 85.5 & 98.3 \\
    \hline
    \end{tabular}%
  \label{tab:apt36k}%
\end{table*}%

\textbf{COCO-W val set.} Different from the previous datasets, COCO-W contains not only the annotations for typical human body keypoints, but also provides fine-grained annotations on the faces, hands, and feet. Similar to the MS COCO val set, we use the person detection results from SimpleBaseline~\cite{xiao2018simple} for evaluation. As shown in Table~\ref{tab:wholebody}, our ViTPose++-B model obtains competitive performance at 57.4 AP on the COCO-W val set. With the model becoming larger, the performance of ViTPose++ further increases, and ViTPose++-H sets a new SOTA, \eg, 61.2 AP, demonstrating the excellent scalability of ViTPose++. 

\textbf{AP-10K val set.} To evaluate the performance of ViTPose++ for animal pose estimation, we adopt the AP-10K and APT-36K datasets. We use an input size of 256$\times$192 for ViTPose++ during training, while other methods take a larger size of 256$\times$256 by default. The results are listed in Table~\ref{tab:ap10k}. As can be seen, ViTPose++ performs well not only for human pose estimation but also for animal pose estimation. For example, ViTPose++-B outperforms previous methods by obtaining 74.5 AP. In comparison, the performance of the largest model ViTPose++-H further increases to 82.4 AP, setting a new record on the animal pose estimation task and demonstrating the potential of ViTPose in dealing with various types of body pose estimation tasks.

\textbf{APT-36K val set.} We also evaluate the performance of ViTPose++ on the APT-36K dataset, which contains a larger number of instances. As shown in Table~\ref{tab:apt36k}, ViTPose++ also achieves better performance on the APT-36K datasets than previous methods. Note that ViTPose++-S delivers a much higher AP on APT-36K than AP-10K, \ie, 74.2 AP v.s. 71.4 AP, probably due to the more balanced data distribution of APT-36K than that of AP-10K.

\begin{figure*}[htbp]
    \centering
\includegraphics[width=0.8\linewidth]{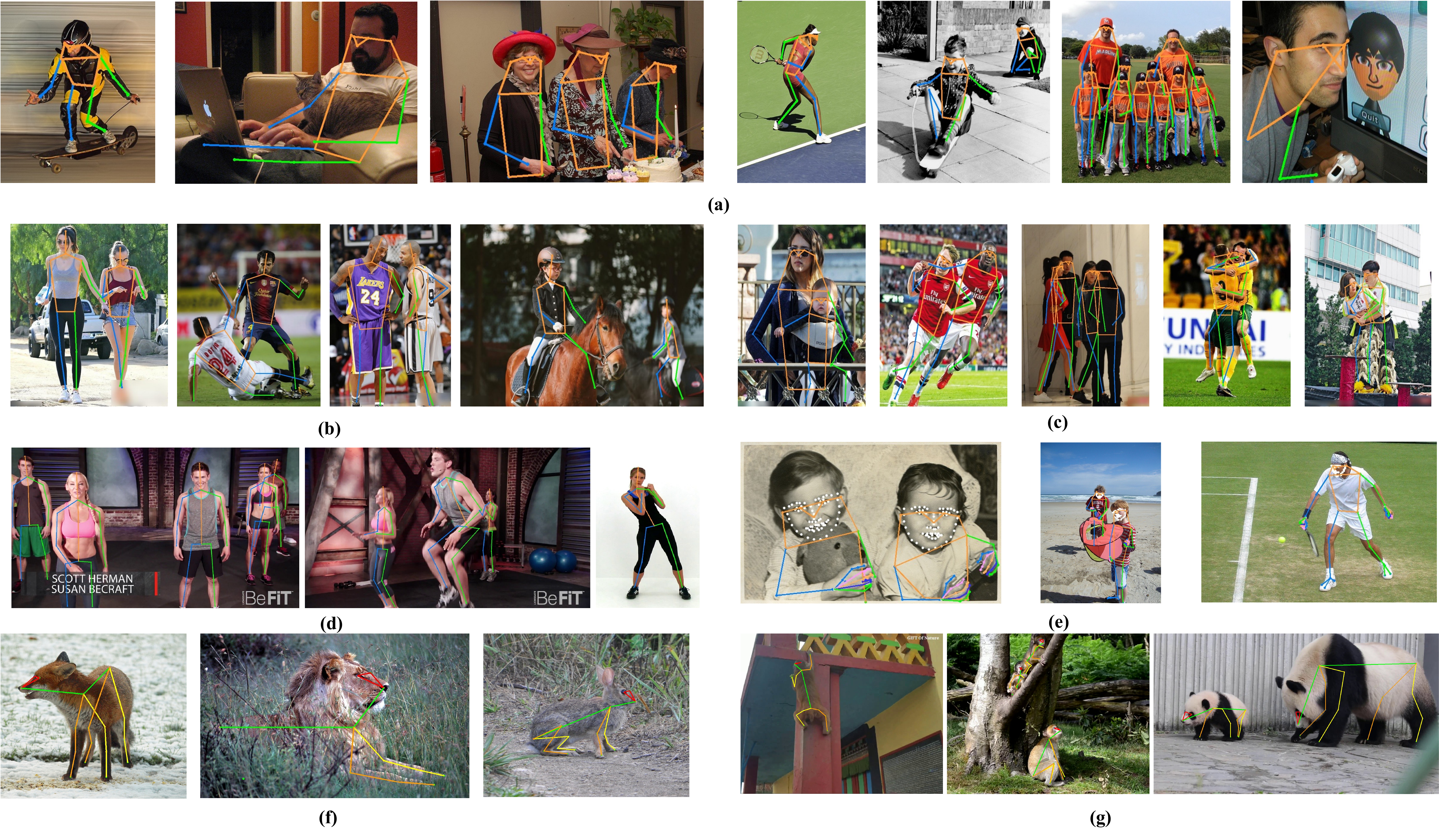}
    \caption{Results of ViTPose++-H on the (a) MS COCO, (b) AIC, (c) OCHuman, (d) MPII, (e) COCO-W, (f) AP-10K, and (g) APT-36K.}
    \label{fig:subjective_one}
\end{figure*}


\subsection{Subjective results}

We also provide some visual pose estimation results for subjective evaluation. {We show the results of ViTPose++-H on the MS COCO, AIC, OCHuman, MPII, COCO-W, AP-10K, and APT-36K datasets, respectively. The results are shown in Fig.~\ref{fig:subjective_one}.} As can be seen, ViTPose++ is good at dealing with challenging cases such as occlusions, blur, scale changes, appearance variance, odd body postures, and complex backgrounds, owing to its strong representation ability and flexibility in encoding pose-relevant knowledge from multiple types of body pose estimation datasets.







\begin{figure*}[htbp]
    \centering
    \includegraphics[width=0.9\linewidth]{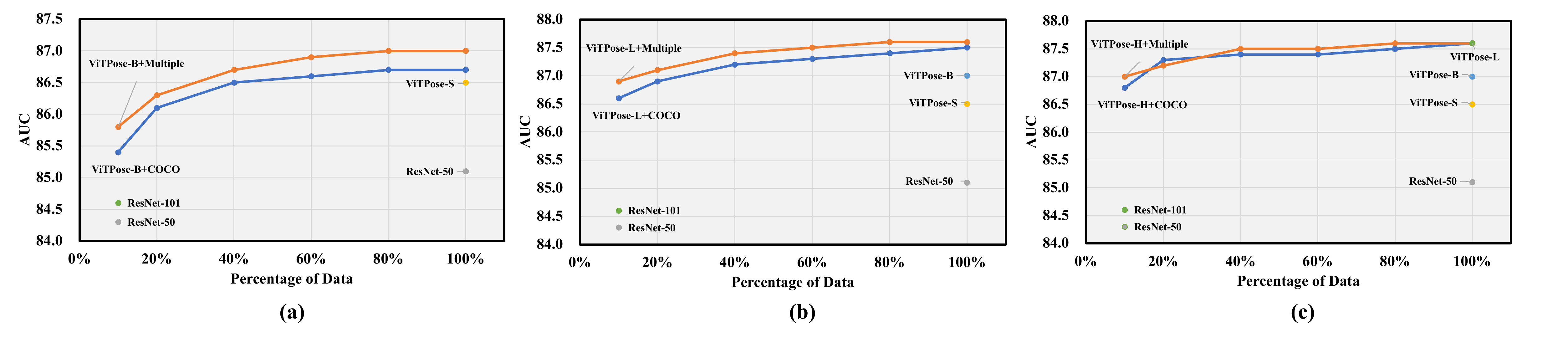}
    \caption{{The transfer learning results of ViTPose-B (a), ViTPose-L (b), and ViTPose-H (c) on the Interhand2.6M dataset using different percentages of training data. The red and blue curves denote the results of ViTPose at the multi-task and single-task training settings. We also plot the results of small pose estimation models, \eg, SimpleBaseline~\cite{xiao2018simple} with ResNet-50 and ResNet-101, for reference.}}
    \label{fig:dataefficiency_hand}
\end{figure*}

\subsection{Data efficiency analysis}
A critical property of the foundation model is high data efficiency for transfer learning, \ie, performing well on a target domain after fine-tuning on only a small fraction of labeled data. To evaluate the data efficiency of ViTPose under different settings, we first train them with only the MS COCO dataset and the combination of MS COCO, MPII, AIC, and COCO-W datasets, respectively. Then, we demonstrate their generalization ability by further fine-tuning them on two different datasets, \ie, InterHand2.6M~\cite{moon2020interhand2} and AP-10K~\cite{yu2021ap}, to make a comprehensive evaluation of the generalization ability on datasets with different domain shifts, \ie, human body $\to$ human hand and human body $\to$ animal body. Specifically, we use the 10\%, 20\%, 40\%, 60\%, 80\%, and 100\% percentage of the training data for fine-tuning. During fine-tuning, we ensure the models are fine-tuned with the same amount of data by keeping the same number of iterations.

\begin{figure}[htbp]
    \centering
    \includegraphics[width=0.7\linewidth]{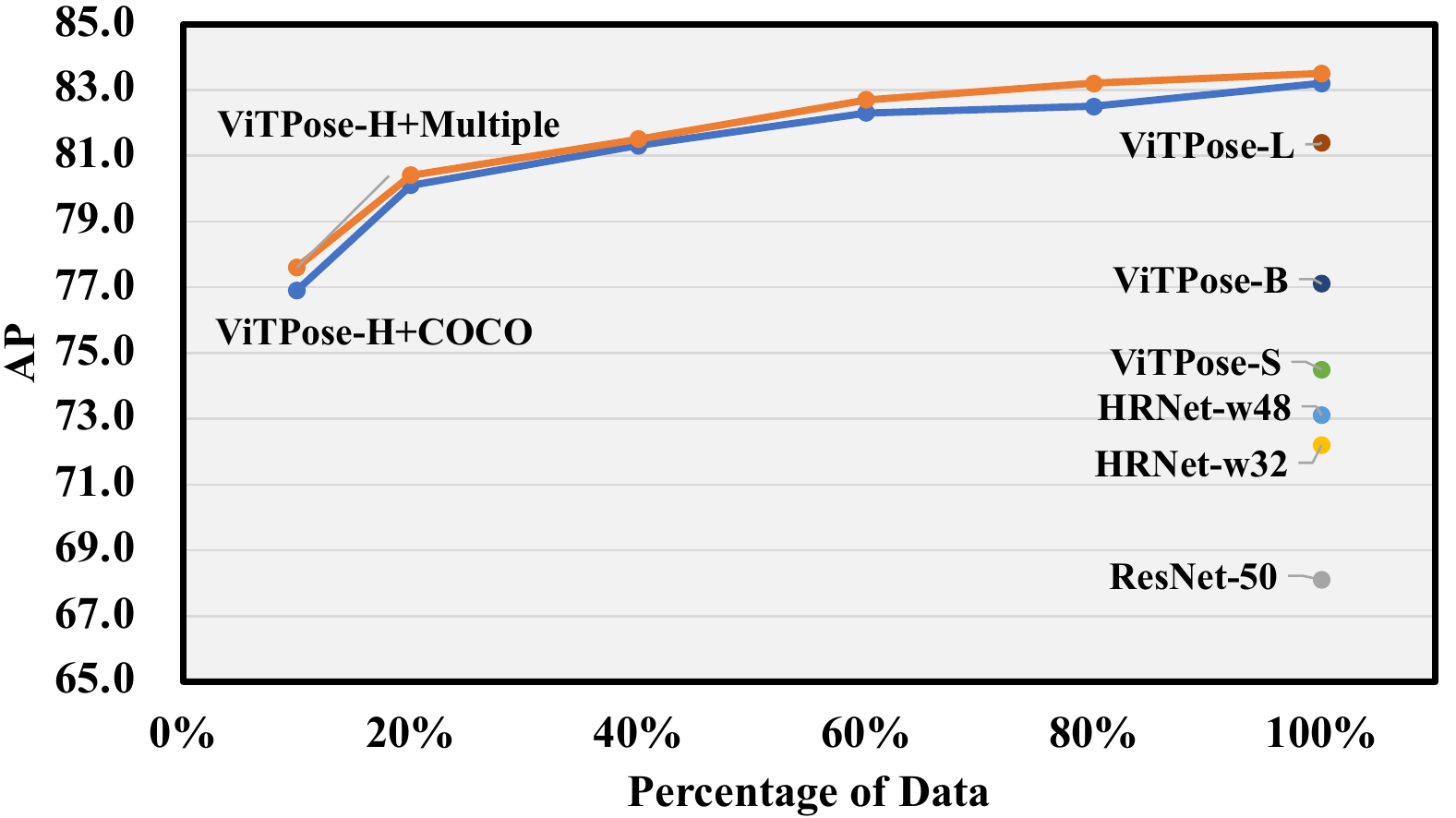}
    \caption{The transfer learning results of of ViTPose-H on the AP-10K~\cite{yu2021ap} using different percentages of training data. We also plot the results of smaller models for comparison.}
    \label{fig:dataefficiency_ap10k}
\end{figure}

\begin{figure*}[htbp]
    \centering
    \includegraphics[width=0.85\linewidth]{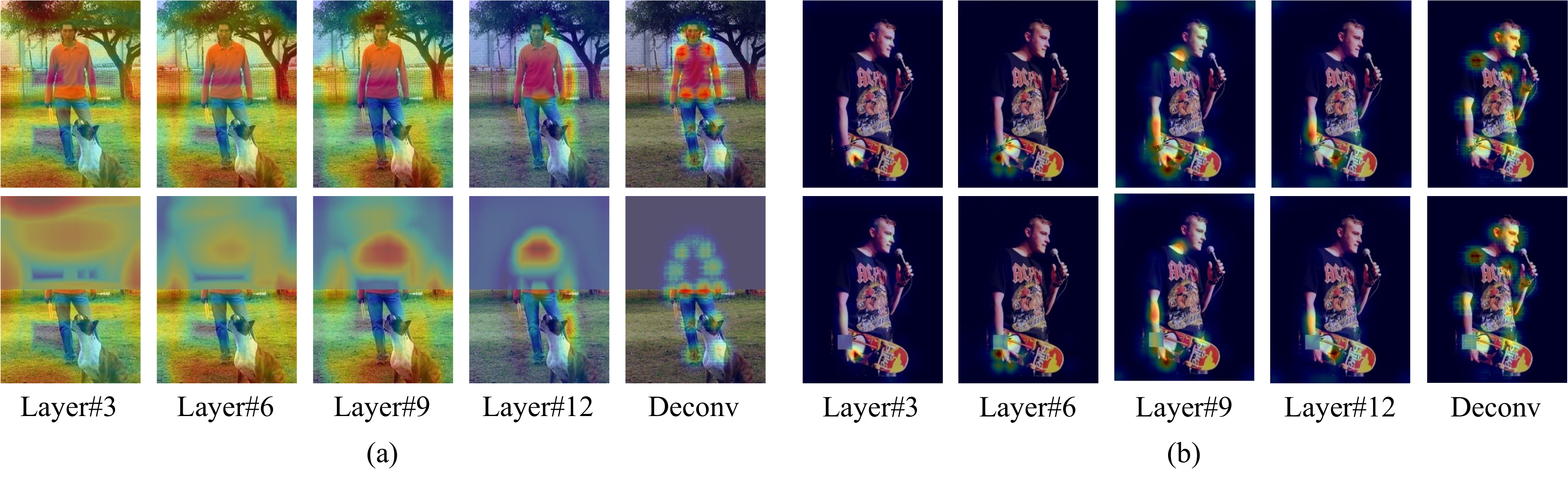}
    \caption{Visualization of the (a) feature maps and (b) attention maps from ViTPose++-B on two test images w/ and w/o manual masks.}
    \label{fig:feature_half}
\end{figure*}

\begin{figure}[htbp]
    \centering
    \includegraphics[width=0.7\linewidth]{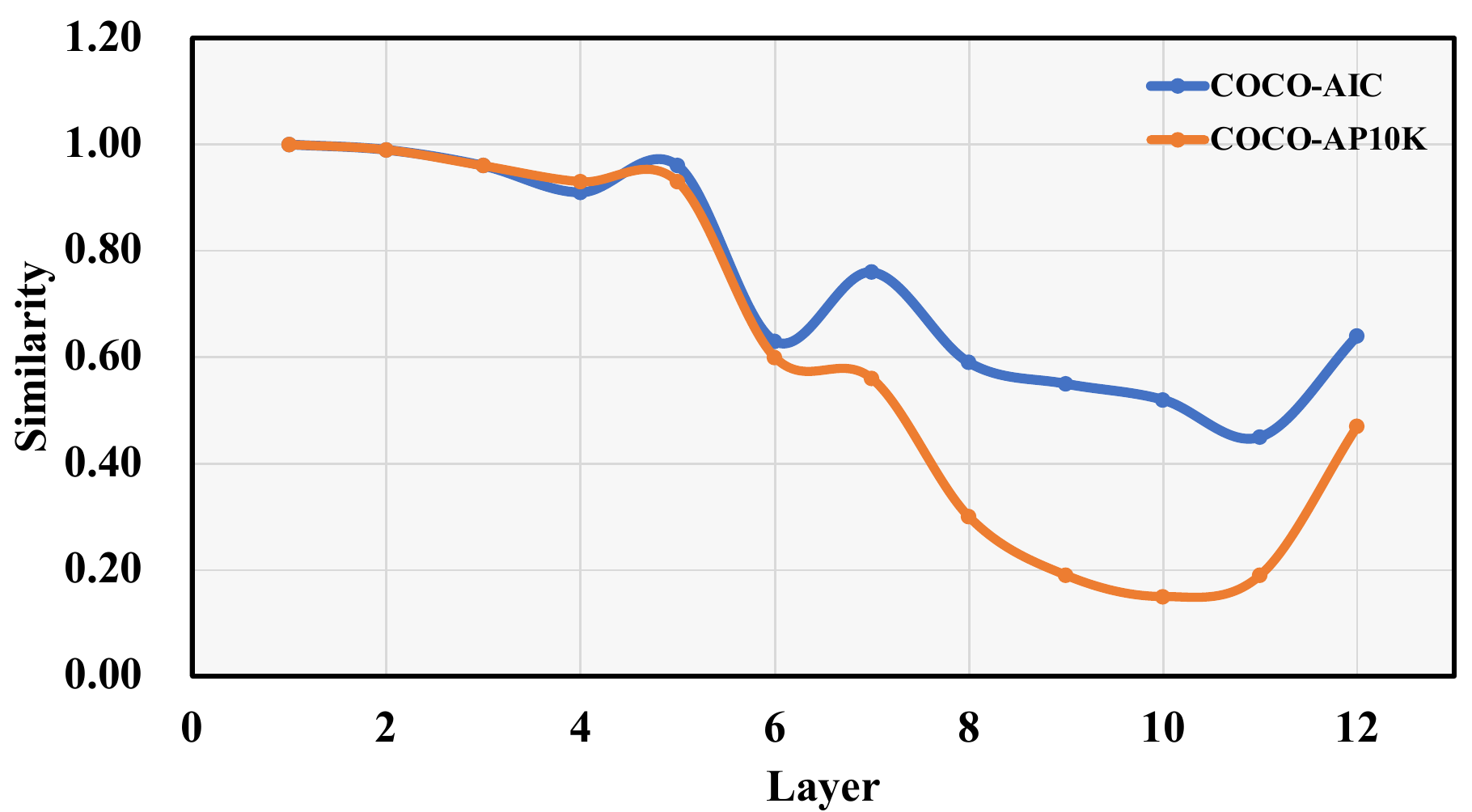}
    \caption{The similarity of task-specific FFN weights of ViTPose++-B, \ie, COCO v.s. AIC and COCO v.s. AP-10K.}
    \label{fig:MoESim}
\end{figure}

\textbf{InterHand2.6M.} We first conduct the data efficiency experiments using ViTPose-B, ViTPose-L, and ViTPose-H on the InterHand2.6M datasets. The results are plotted in Fig.~\ref{fig:dataefficiency_hand}, where we also provide the results of representative small models trained with 100\% training data for reference. It can be observed that the data efficiency is improved with the increase of the model size, \ie, with the same percentage of data for training. ViTPose with the ViT-H backbone always has better performance than that with smaller ones. {Moreover, the transformer-based methods obtain better performance gains compared with the CNN-based method, \ie, increasing the transfer data from 10\% to 100\%, SimpleBaseline with a ResNet-50 backbone only experiences 0.8 AUC increase while ViTPose-B obtains 1.3 AUC improvement.} Besides, multi-task training on multiple types of body pose estimation datasets also improve data efficiency. For example, compared with single-task training, the multi-task training helps ViTPose-B/L/H achieve a gain of 0.4/0.3/0.2 AUC when using 10\% data for training. In addition, compared with the small models using 100\% data for training, ViTPose-H obtains better performance with fewer training data, \eg, ViTPose-H using 20\% training data (87.2 AUC) outperforms ViTPose-S (86.5 AUC) and SimpleBaseline with ResNet-50 (85.1 AUC) using 100\% training data, showing the high data efficiency of large models. Besides, with the model size increasing, the benefit of using multiple types of pre-training data becomes lower, \ie, the gap between the red and blue curves becomes narrower. It implies that the model size plays a more important role in improving the data efficiency of transfer learning, \ie, suggesting the notion that \textit{more is different}.

\textbf{AP-10K.} Similarly, we evaluate the data efficiency of the previous best model ViTPose-H on the AP-10K dataset. The results are shown in Fig.~\ref{fig:dataefficiency_ap10k}. It can be observed that with only 10\% data for training, ViTPose-H outperforms the representative methods with small models, \eg, ResNet-50, HRNet-32, and HRNet-48, as well as the ViTPose-B and ViTPose-S models, further validating the good data efficiency of the large model.



\subsection{Visualization and analysis} 
\textbf{Visualization of feature maps.} To inspect the learning ability of ViTPose, we visualize the feature maps of ViTPose++-B on the normal and heavily occluded images, \ie, we mask half of the images. As shown in Fig.~\ref{fig:feature_half} (a), ViTPose++-B gradually focuses on the human body from the shallow layers to deep layers in the backbone network. The up-sampling layer further mitigates the interference of the background and focuses on the keypoint locations. When comparing the feature maps extracted from the normal images (1st row) and masked images (2nd row), the features are almost the same in the visible parts. In contrast, ViTPose gradually guesses the locations of those invisible joints for the masked parts. For example, although the heads of the human are masked, ViTPose can still guess the locations of the head and the keypoints based on the visible parts, as demonstrated in the upper parts of the feature maps from the 9th and 12th layers. 

\textbf{Visualization of attention maps} To further inspect how ViTPose models the relationships between different keypoints, we visualize the attention maps regarding the right wrist from the 3rd, 6th, 9th, and 12th layers of ViTPose++-B. We test on both normal images (1st row) and images with the right wrist masked (2nd row). The results are shown in Fig.~\ref{fig:feature_half} (b). It can be observed that the attention map is rather similar no matter whether the right wrist is masked or not, especially in the deep layers. For example, in the 12th layer, ViTPose pays attention to the person's right arm to help localize the locations of the right wrist in both cases. These visualization results present an intuitive explanation of how ViTPose leverages the learned relationship between different keypoints to facilitate the prediction of a specific (invisible) keypoint.

\textbf{Task correlation.} We further evaluate the correlation between different types of body pose estimation tasks by calculating the similarity between the weights of the task-specific FFN parts. {The weight similarity is defined as the cosine similarity between the weights of the FFN layer corresponding to each dataset, respectively. We analyze the per-layer similarity of ViTPose++-B model.} The results are plotted in Fig.~\ref{fig:MoESim}. It can be observed that the task-specific FFNs for human pose estimation have a larger similarity than that of the task-specific FFNs for body pose estimation of different species, \ie, the similarity between MS COCO and AIC is larger than that between MS COCO and AP-10K, especially in the deeper layers, implying that there probably exist conflicts between different types of body pose estimation tasks. Nevertheless, for the shallow layers, the weights of FFN layers are rather similar in both cases, implying that FFNs in the deeper layers are more task-specific than those in the shallower layers, and thus we may only adopt MoE in the deeper FFN layers.

{
\begin{figure}
    \centering
    \includegraphics[width=0.9\linewidth]{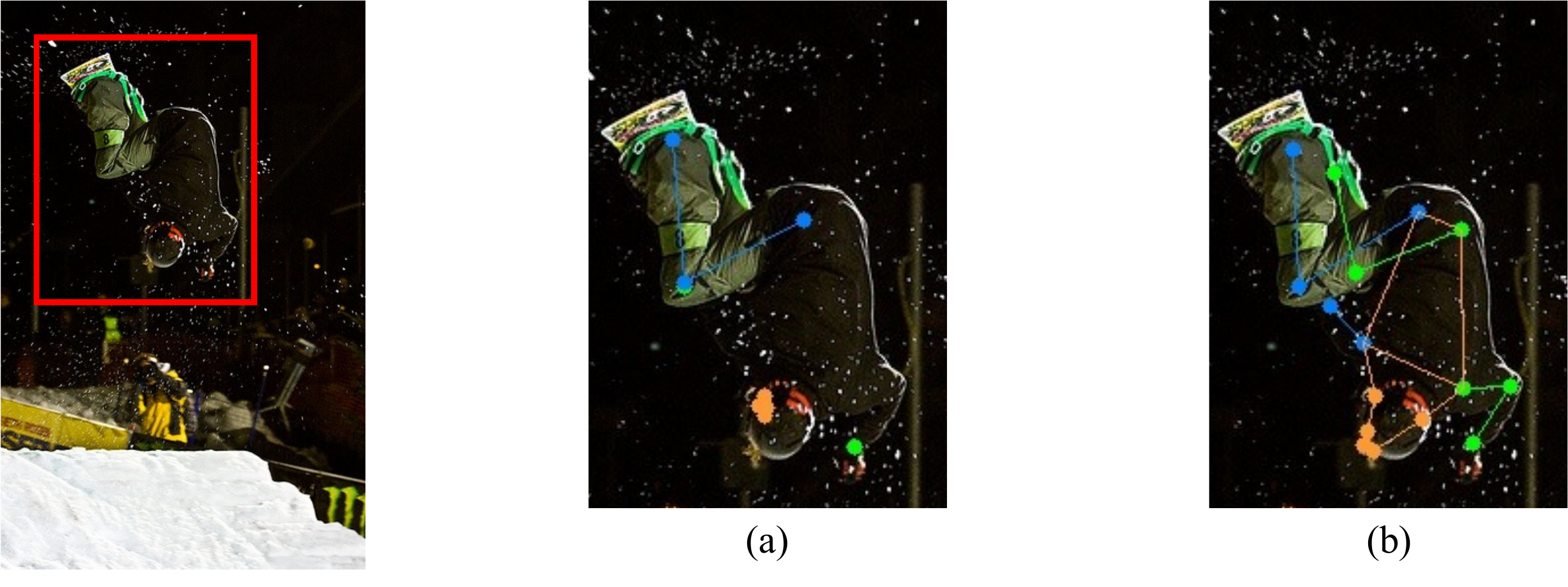}
    \caption{Failure case analysis on a test image with an extreme pose. (a) Result of ViTPose-S. (b) Results of ViTPose++-H.}
    \label{fig:failure}
\end{figure}

\subsection{Failure case analysis}
Despite the superior performance of ViTPose on various pose estimation tasks, there are still challenges when it comes to handling extreme postures, such as skiing as in Fig.~\ref{fig:failure}. Fortunately, due to the scalability of ViTPose, these limitations can be mitigated by using a larger backbone and incorporating more data during training. For instance, the ViTPose++-H model, trained with an expanded dataset containing a greater variety of pose data, can estimate the pose correctly. 
}

\section{Limitations and discussion}
\label{sec:limit}
Despite the good properties and performance of our ViTPose and ViTPose++ for body pose estimation, even without elaborate structural designs, there is still room to improve them further. First, we only focus on the pose estimation task in this study, while designing a unified foundation model for simultaneous object detection (tracking) and pose estimation is more appealing for generic body pose estimation and tracking. Besides, we only use the training data in the visual modality in this study, while leveraging language knowledge of body keypoints to help body pose estimation, especially for zero-shot generalization on unseen species or keypoints, is also worth further exploration. 

\section{Conclusion}
{This paper presents ViTPose and ViTPose++ as the simple baseline for body pose estimation.} They demonstrate good properties, including simplicity, scalability, flexibility, and transferability, which have been well justified through extensive experiments on the representative benchmarks, including MS COCO, AIC, MPII, OCHuman, COCO-W, AP-10K, and APT-36K. New performance records on these datasets have also been set by the proposed models. {We hope this work could provide useful insights to the community and inspire more future studies on the potential of developing vision transformers towards a foundation model for generic pose estimation and beyond.}

\ifCLASSOPTIONcaptionsoff
  \newpage
\fi

\bibliographystyle{IEEEtran}
\bibliography{egbib}

\begin{IEEEbiography}[{\includegraphics[width=1in,height=1.25in,clip,keepaspectratio]{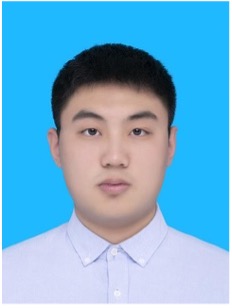}}]{Yufei Xu} received the B.E. degree from the University of Science and Technology of China (USTC), Hefei, China, in 2019. He is currently pursuing the Ph.D. degree with The University of Sydney, under the supervision of Prof. Dacheng Tao and Dr. Jing Zhang. His research interests include computer vision and deep learning. \end{IEEEbiography}

\begin{IEEEbiography}[{\includegraphics[width=1in,height=1.25in,clip,keepaspectratio]{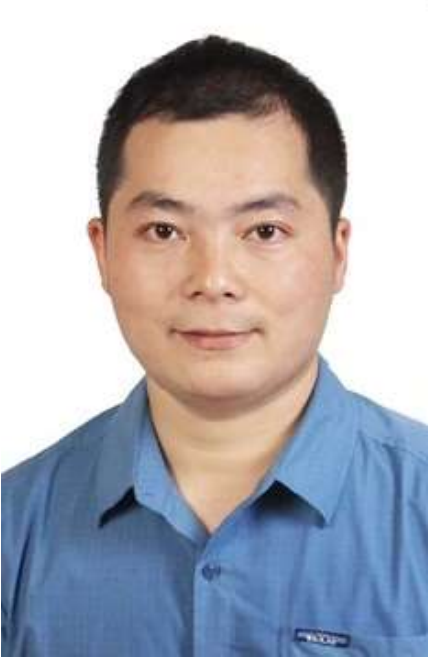}}]{Jing Zhang} (Senior Member, IEEE) is currently a Research Fellow at the School of Computer Science, The University of Sydney. He has published more than 60 papers in prestigious conferences and journals, such as CVPR, ICCV, ECCV, NeurlPS, ICLR, IEEE TPAMI, and IJCV. His research interests include computer vision and deep learning. He is also a Senior Program Committee Member of the AAAI Conference on Artificial Intelligence and the International Joint Conference on Artificial Intelligence. He serves as a reviewer for many prestigious journals and conferences. \end{IEEEbiography}

\begin{IEEEbiography}[{\includegraphics[width=1in,height=1.25in,clip,keepaspectratio]{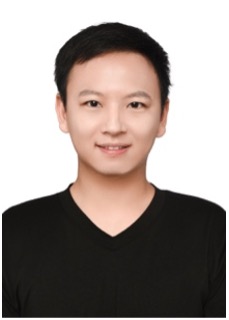}}]{Qiming Zhang} is currently pursuing the Ph.D. degree with the School of Computer Science, the University of Sydney under the supervision of Prof. Dacheng Tao and Dr. Jing Zhang. He received the B.Sc. degree from Zhejiang University in 2017, and the M.Phil degree from the University of Sydney in 2019. His research interests include transfer learning, recognition, and vision transformer in computer vision. \end{IEEEbiography}

\begin{IEEEbiography}[{\includegraphics[width=1in,height=1.25in,clip,keepaspectratio]{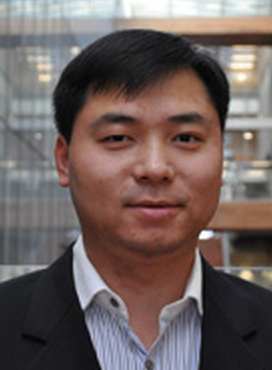}}]{Dacheng Tao} (F'15) is currently a Professor of Computer Science, Peter Nicol Russell Chair and an Australian Laureate Fellow in the Sydney AI Centre and the School of Computer Science in the Faculty of Engineering at The University of Sydney. He was the founding director of the Sydney AI Centre. He mainly applies statistics and mathematics to artificial intelligence and data science, and his research is detailed in one monograph and over 200 publications in prestigious journals and proceedings at leading conferences. He received the 2015 and 2020 Australian Eureka Prize, the 2018 IEEE ICDM Research Contributions Award, and the 2021 IEEE Computer Society McCluskey Technical Achievement Award. He is a Fellow of the Australian Academy of Science, AAAS, ACM and IEEE.\end{IEEEbiography}

\end{document}